\documentclass[10pt,letterpaper]{article}
\PassOptionsToPackage{numbers,sort&compress}{natbib}
\usepackage[preprint]{neurips_2026}
\usepackage[utf8]{inputenc}
\usepackage[T1]{fontenc}

\usepackage[dvipsnames]{xcolor}
\usepackage{graphicx}
\usepackage{amssymb}
\usepackage{booktabs}
\usepackage{subcaption}
\usepackage[shortlabels,inline]{enumitem}
\usepackage[hyphens]{url}
\usepackage{fvextra}
\DefineVerbatimEnvironment{verbatim}{Verbatim}{fontsize=\small,breaklines=true}
\usepackage{newunicodechar}
\newunicodechar{“}{``}
\newunicodechar{”}{''}
\newunicodechar{’}{'}
\newunicodechar{–}{--}
\newunicodechar{—}{---}
\usepackage{nicefrac}
\usepackage{pmboxdraw}
\usepackage{array}
\usepackage{mathtools}
\usepackage{microtype}
\usepackage[disable]{todonotes}
\usepackage{glossaries}

\usepackage{amsmath,amsfonts,bm}

\def\eqref#1{equation~\ref{#1}}

\def\1{\bm{1}}

\DeclareMathAlphabet{\mathsfit}{\encodingdefault}{\sfdefault}{m}{sl}
\SetMathAlphabet{\mathsfit}{bold}{\encodingdefault}{\sfdefault}{bx}{n}

\DeclareMathOperator*{\argmin}{arg\,min}

\definecolor{linkblue}{rgb}{0.21,0.49,0.74}
\usepackage[breaklinks,colorlinks,allcolors=linkblue]{hyperref}
\usepackage[capitalize,noabbrev]{cleveref}
\hypersetup{
    pdfauthor={Melanie F. Pradier, Camilla Longden, Anja Thieme, Rita F. Marques, Nari Johnson, Sidhika Balachandar, Xinnuo Xu, Linda Y. Wen, Martin Grayson, Katja Hofmann, Cecily Morrison, Daniela Massiceti},
    pdftitle={Measuring Preferred Representation in AI-generated Images with Community Scorers}
}
\newacronym{mllm}{MLLM}{Multimodal Large Language Model}
\newacronym{blv}{BLV}{Blind or Low-Vision}
\newacronym{dwf-s}{DWF-S}{Dwarfism for Global South}
\newacronym{dwf-n}{DWF-N}{Dwarfism for Global North}

\newcolumntype{C}[1]{>{\centering\arraybackslash}p{#1}}   

\newcolumntype{P}[1]{>{\centering\arraybackslash}p{#1}}

\crefname{section}{Sec.}{Secs.}
\Crefname{section}{Sec.}{Secs.} 
\crefname{figure}{Fig.}{Figs.}
\Crefname{figure}{Fig.}{Figs.}
\crefname{table}{Tab.}{Tabs.}
\Crefname{table}{Tab.}{Tabs.}

\title{Measuring Preferred Representation in AI-generated Images with Community Scorers}
\author{%
\textbf{Melanie F.~Pradier}$^{1,}$\thanks{Correspondence to: melanief@microsoft.com, cecilym@microsoft.com}
\quad
\textbf{Camilla Longden}$^{1}$
\quad
\textbf{Anja Thieme}$^{1}$
\\
\textbf{Rita F. Marques}$^{1}$
\quad
\textbf{Nari Johnson}$^{1,3}$
\quad
\textbf{Sidhika Balachandar}$^{1,4}$
\\
\textbf{Xinnuo Xu}$^{1}$
\quad
\textbf{Linda Y. Wen}$^{1}$
\quad
\textbf{Martin Grayson}$^{1}$
\\
\textbf{Katja Hofmann}$^{1}$
\quad
\textbf{Cecily Morrison}$^{1,\star}$
\quad
\textbf{Daniela Massiceti}$^{1,2}$
\\[0.3cm]
$^{1}$Microsoft Research, Cambridge, United Kingdom\\
$^{2}$Canva Research, Sydney, Australia\\
$^{3}$Carnegie Mellon University; Pittsburgh, PA USA\\
$^{4}$University of California; Berkeley, CA USA
}

\begin{document}
\maketitle
\begin{abstract}
AI-generated images often misrepresent historically marginalised communities, with real-world consequences beyond the images themselves.
However, conventional metrics fail to measure how communities want to be represented in AI-generated images. To address this, we introduce the concept of \textit{preferred representation}. Building on previous work, we propose a measurement instrument that operationalises this concept through a \textit{Community Scorer} -- a model that evaluates a specific community's preferred representation in AI-generated images. We interrogate the validity of Community Scorers on three diverse disability communities that span the Global North and South. The results show that Community Scorers align significantly better with community ratings compared to other widely used text-to-image metrics. We then demonstrate the usefulness of these scorers in a downstream content filtering task. We release the Community Preference Dataset to enable others to optimise how Community Scorers can be scaled for culturally specific evaluation of AI-generated visual media.
\end{abstract}
\section{Introduction}
\label{sec:intro}

Visual media plays a pivotal role in shaping public perception of historically marginalised communities \citep{ellis2020routledge}. It can embed stereotypes or drive positive change by shifting public perspectives \citep{photogpt2025visualrevolution} in ways that increase access to livelihoods, such as education and employment. Unfortunately, AI image generation consistently underperforms for marginalised communities, replicating biases \citep{ghosh2024do}, reinforcing stereotypes \citep{bianchi2023easily, jha2024visage}, and erasing key cultural identity markers \citep{hall2024towards}. Persistent failures to represent communities as they prefer can leave them vulnerable to broad societal consequences.

Effective, scalable evaluation is needed to address these misrepresentations \citep{biega2025towards}. Yet, assessing how well AI-generated images represent communities using current automated measurement approaches is challenging. Most automated metrics prioritise image--prompt alignment, realism, or aesthetics which do not directly assess how communities desire to be represented \citep{kannen2024beyond, nayak2025culturalframes, hall2024towards}. Other work focuses on isolated issues, such as anatomical errors \citep{gandikota2024concept, li2025aghi, lu2024handrefiner}, demographic biases such as skin tone or gender \citep{friedrich2025auditing, huangwherefact2025, naik2023social}, or correctness of cultural artefacts \citep{basu2023inspecting, hall2023dig, magomere2024you, ventura2025navigating}. Yet, research suggests that these measurements may fail to capture people's preferences, which often move beyond a single dimension to reflect the rich `social world' of their culture~\citep{qadri2023ai}.

A particular challenge of measuring cultural representation at scale is its plurality \emph{across} different communities \citep{chasalow2021representativeness}. While attempts have been made to synthesise key elements of culture \citep{adilazuarda2024towards, qadri2025case}, it is acknowledged that what is evaluated must be community-specific \citep{mack2024they, mim2024between}. Recent work calls for community participation to support pluralistic evaluation \citep{qadri2025case, rastogi2025whose}, but research that centres community members as both knowledge bearers and decision makers in developing scalable metrics is still nascent \citep{johnson2026evaluating}.

\begin{figure*}
    \centering
    \includegraphics[width=1\linewidth]{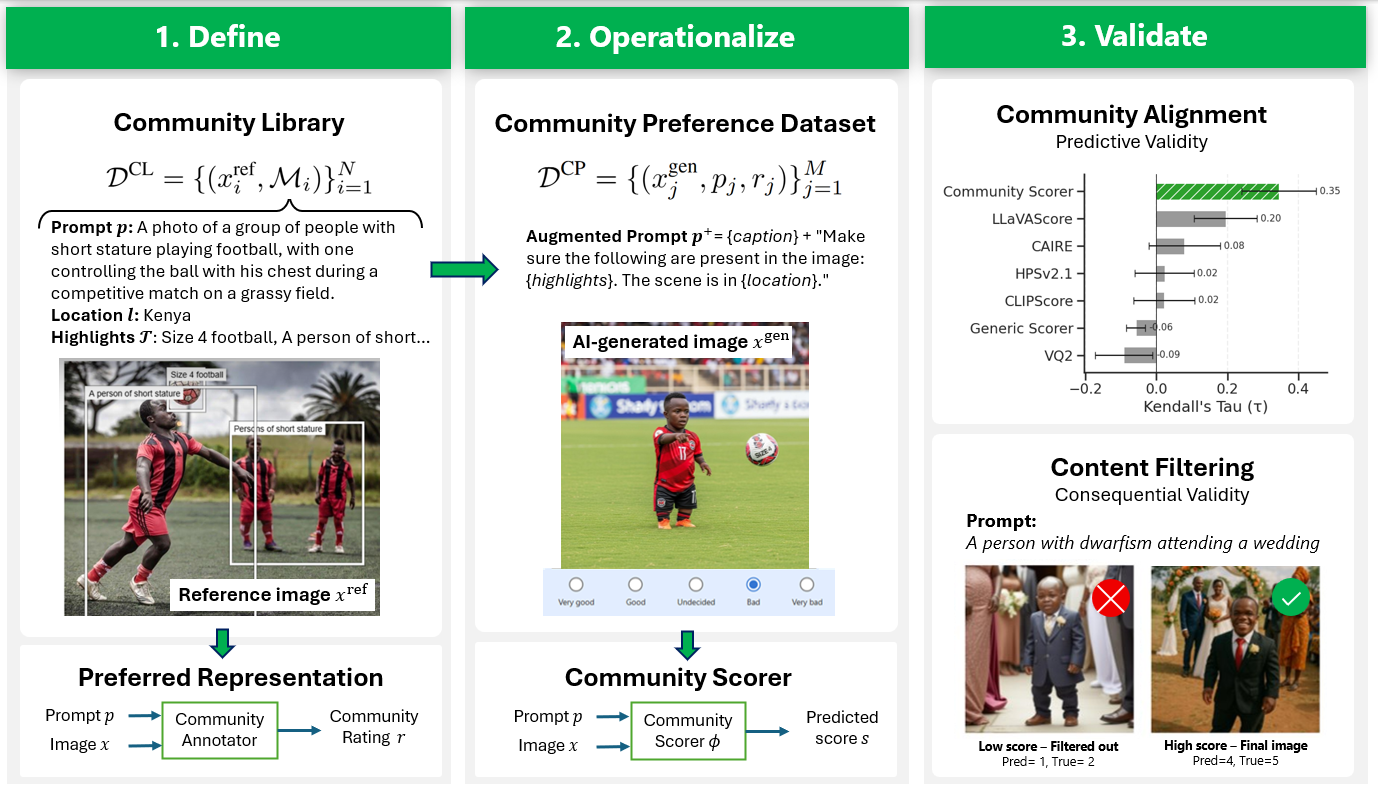}
    \caption{\textbf{Our contributions.} We define, operationalise, and validate an automated metric of preferred representation. The Community Library corresponds to $\mathcal{D}^{\text{CL}}$ and the Community Preference Dataset to $\mathcal{D}^{\text{CP}}$. The Preferred Representation box (bottom left) represents the target function we seek to approximate via a scalable proxy (bottom middle), the Community Scorer $\phi$.}
    \label{fig1:motivation}
\end{figure*}

We address this gap by formalising the concept of `preferred representation' -- the way a community wishes to be visually represented in AI-generated images. We lean on the framework of \cite{adcock2001measurement, chouldechova2024shared}, and draw apart the \emph{systematisation} of the concept -- how it is defined -- from its \emph{operationalisation} -- how it is measured. We build on prior work \citep{thieme2026engaging} in which communities systematise their preferred representation through the creation of a Community Library. This approach scaffolds a member community (such as a non-profit organisation) to jointly deliberate and formulate their representation and then express that through the collection and curation of carefully selected, richly annotated photographs.

We use this Community Library to operationalise a Community Scorer, an automated measurement instrument for how well a new AI-generated image aligns with a specific community's preferred representation. 
We first build the Community Preference Dataset using AI-generated images based on text prompts from the Community Library. The dataset contains ratings by the designated community lead of how well these images represent the community.
We then train the Community Scorer and interrogate its validity. Figure~\ref{fig1:motivation} provides an overview and worked example of this process: 1) Define; 2) Operationalise; and 3) Validate.

We develop Community Scorers for three globally diverse disability communities. We then validate them by demonstrating significantly better alignment with community member ratings of AI-generated images than conventional automated metrics (e.g.~CLIPScore) or off-the-shelf multimodal large language models (MLLMs). We also demonstrate their usefulness in a downstream content filtering task. In summary, our contributions are as follows:
\begin{enumerate}[leftmargin=*, nosep]
    \item We introduce the measurement construct `preferred representation' for evaluating AI-generated images, with the intention of providing communities with a way to define a scalable metric of how they would like to be represented.
    \item We propose a method to operationalise `preferred representation' via Community Scorers: evaluator models trained on community ratings of AI-generated images that enable scalable, automated assessment of \mbox{preferred representation}.
    \item We demonstrate the creation and validation of Community Scorers with three globally diverse disability communities, showing their potential to align significantly better with community ratings than baselines, as well as their success in a downstream content filtering task.
\end{enumerate}

\section{Related Literature}
\label{sec:related_work}

\paragraph{Metric development in AI research.}
A growing body of work calls for greater rigour in the development of evaluations \citep{saxon2024who, wallach2025position}. In this paper, we draw upon \cite{adcock2001measurement, chouldechova2024shared}, which has translated work from measurement theory in the social sciences for machine learning evaluation. A methodological principle underpinning this work is the differentiation between the definition of a concept to be measured -- its \emph{systematisation} -- and the actual measure itself -- its \emph{operationalisation}. They further argue for the need to validate the operationalised measure with respect to the clearly systematised concept. We follow this approach, separating these two stages and focusing on the validated \emph{operationalisation} of the concept of ``preferred representation''.

As the field gets more precise about measures, literature has delineated many of the ways that common measurement approaches have elided cultural differences \citep{munir2026consensus, wan2023everyone}. In response, machine learning advances have focused on developing pluralistic alignment strategies (e.g.~\citealp{halpern2025pairwise}). Other work has focused on preserving individual differences in reward models rather than collapsing heterogeneous preferences into a single aggregate signal \citep{barreto2025capturing}. These approaches are an important way of preserving cultural difference in a single generalised model. In this paper, we consider an alternative way to preserve plurality. We develop per-community evaluators rather than attempting to create a model that can generalise across all communities.

\paragraph{Participation in metric development.}
Methods for public involvement and participation that account for the particular socio-technical properties of AI are not well developed \citep{modhvadia2024participatory}. \citet{nguyen2026validating} pose that a particular advantage of the separation between systematisation and operationalisation is the explicit opportunity to bring stakeholders' voices into the measurement process in the form of defining, or systematising, the concept of measurement. \citet{suresh2024participation}, looking more broadly at participation in the development of AI systems, put emphasis on communities shaping an AI system for a specific purpose at the per-community level, rather than a focus on the frontier model itself. Building on these ideas, we invite community members to participate in designing per-community evaluation metrics that capture the preferred representation that they had previously systematised.

Involving communities (and their membership) in measurement has been limited. Most existing human-centred approaches involve communities in defining evaluation datasets, e.g.~\citep{hall2025human, seth2024dosa, hamna2026building, rastogi2026places}, annotating data \citep{hada2024akal}, or manually evaluating produced images \citep{bennett2025toward, gillespie2026ai, mack2024they, qadri2023ai}. More recently, \cite{johnson2026evaluating} have attempted to scale community participation through automated (M)LLM judges by developing text-based rubrics for evaluating AI-generated images of cultural artefacts based on interviews with community members. This work highlights the complexity and iteration required in translating community knowledge into scalable metrics, often requiring researchers to play critical bridging roles which limits scalability of these methods to many communities.

A key part of measurement development is the annotation of examples used as ground truth. Research has clearly shown that who annotates matters \citep{wan2023everyone}, with annotations often differing along common demographic lines. Research has considered what conditions make annotation a participatory activity, reframing it as the activities that give authority over problem framing (e.g.~definition of categories). Annotation tasks where multidimensional rating is performed, for example, would not be considered participatory if the dimensions were defined by researchers \citep{diebold2026participatory}. \citet{coyne2026three} provides another perspective, defining annotation as extension, evidence, or authority. In the last model, a person has been given authority to speak on behalf of a group of people. This is the model that we use in working with a designated community lead.

\paragraph{Community libraries.}
Community Libraries \citep{thieme2026engaging} engage communities in creating exemplar-based systematisations that can be directly operationalised. Community organisations, as stewards of a community's collective voice \citep{larana1994new}, are provided with a structured workflow within an online data stewardship platform (see appendix). A designated community lead works with community members to define their `preferred representation'.

Each resulting Community Library consists of 400 real photographs structured by 3--5 named, emergent themes (e.g., education, employment) and sub-themes (e.g., blind teachers, students with visual impairment in primary school) that are meaningful to the community. Each image within the library has a detailed text prompt that describes the salient elements of preferred representation. Images also have bounding box labels that specify human or material elements important to cultural representation (see Figure~\ref{fig1:motivation}). Taken together, these exemplars implicitly define (i) the dimensions along which representations vary, and (ii) the breadth and depth of themes -- and their boundaries -- as understood by the community. This approach to systematisation allows representation to be holistically perceived, rather than requiring decomposition into checklists of explicit attributes \citep{kiden2025back}.

\section{Defining Preferred Representation}
\label{sec:systematisation}
We define \textit{preferred representation} as the way a community wishes to be portrayed in visual media. We take the view that representation is deeply contextual and should be defined by and negotiated within the community~\citep{qadri2025confusing}. Most prior work on evaluating representation in AI-generated images focuses on identifying errors or failures, often through the lens of AI safety or bias mitigation~\citep{mack2024they, tevissen2024disability}. While mitigating errors is important, representation is more than just the absence of errors: it encompasses cultural context, social norms, and visual features that express the community's identity and values. For example, for the blindness community in Kenya, not only is it important that a braille display be generated correctly, but that the images show blind students accessing education and careers. This representation helps counter common misconceptions that educating young blind people in Kenya is unimportant due to limited job prospects. Our approach centres on this holistic, positive framing.

As proposed in~\citet{thieme2026engaging}, a community \textit{systematises} preferred representation by curating a Community Library -- a set of $N$ real photographs of themselves paired with structured metadata. We formalise this as:
\begin{equation}
    \mathcal{D}^{\text{CL}} = \{(x_i^{\text{ref}}, \mathcal{M}_i)\}_{i=1}^{N},
\end{equation}
where $x_i^{\text{ref}}$ is a photograph of the community and $\mathcal{M}_i$ is the metadata bundle associated with it. In~\citet{thieme2026engaging}, this bundle includes a text prompt $p_i$ describing the photograph, a set of tags $\mathcal{T}_i$ of salient visual elements in the photo, and the community's geographic region $l_i$. Together, the community library $\mathcal{D}^{\text{CL}}$ visually encodes an organisation-stewarded definition of preferred representation. 

\section{Operationalising Preferred Representation}
\label{sec:operationalisation}

Our goal is to develop a measurement instrument that quantifies how well AI-generated images of a community align with their definition of preferred representation -- the process of \emph{operationalising} the systematised concept. Following the common formulation of image evaluation as scoring image--prompt pairs, we wish to learn a scoring function $\phi:\mathcal{X}\times\mathcal{P}\rightarrow\mathcal{S}$ that maps an AI-generated image $x \in \mathcal{X}$ and its corresponding text prompt $p \in \mathcal{P}$ to a score $s \in \mathcal{S}$ that approximates how a given community would rate that image's alignment with their preferred representation. In our setting, this score approximates how a given community would rate the image's alignment with its preferred representation. We refer to $\phi$ as a Community Scorer. Importantly, $\phi$ is constructed \emph{per community} --- a pluralistic design that preserves each community's distinct, collectively deliberated standards rather than collapsing them into a single normative~model.

Although the Community Library $\mathcal{D}^{\text{CL}}$ captures what preferred representation looks like in real photographs, $\phi$ must ultimately score \emph{AI-generated} images, which are distributionally distinct. We therefore construct a Community Preference Dataset $\mathcal{D}^{\text{CP}}$ of AI-generated images of the given community, derived from $\mathcal{D}^{\text{CL}}$ and rated by the designated community lead on how well they align with the community's preferred representation. This anchors their $\mathcal{D}^{\text{CP}}$ in the same representational space as their $\mathcal{D}^{\text{CL}}$ while exposing $\phi$ to the distribution of images it must score at test time. Next, we describe how we construct $\mathcal{D}^{\text{CP}}$ from $\mathcal{D}^{\text{CL}}$ (Section~\ref{sec:operationalisation-dataset}) and train a Community Scorer $\phi$ (Section~\ref{sec:scorer}).

\subsection{Constructing the Community Preference~Dataset}
\label{sec:operationalisation-dataset}

For each sampled Community Library entry $i$, we form an augmented text prompt $p_i^{+}$ by concatenating its text prompt $p_i$, set of visual tags $\mathcal{T}_i$, and region $l_i$, and use it to produce a set of one or more AI-generated images $\{x_j^{\text{gen}}\}$ of the community. Constructing prompts from $\mathcal{D}^{\text{CL}}$ metadata grounds the AI-generated images in the contexts that the community has defined as important to their preferred representation. In other words, the prompts serve as the link between systematisation and operationalisation.

To train a scorer that reliably differentiates good from bad representations, $\mathcal{D}^{\text{CP}}$ must also span the full quality range of AI-generated images. We therefore select a set of image generation models known to vary in output quality, and generate one image per (prompt, model) pair. In this work, we chose Imagen-4-Ultra \citep{google_imagen4_ultra_2025, saharia2022photorealistic}, GPT-Image-1 \citep{openai_gpt_image_1_2025}, and Stable Diffusion 3.5 Large Turbo \citep{rombach2022highresolution, stabilityai_sd35_large_turbo_2024}, a subset of key models that were available at the time this research was conducted, but any models could be used.

We sampled 100 prompts randomly from $\mathcal{D}^{\text{CL}}$, uniformly distributed among community themes, and created an image from each model to create $M = 300$ samples per community. This balanced the minimum amount of data likely needed to train a scorer against the effort required of the designated community lead. We retrospectively validated this choice with ablation curves: scorer performance converges around 300 ratings. In one community that did not provide this full amount, scorer performance was unreliable (see Section~\ref{sec:validation-agreement}).

We then ask the designated community lead to rate each AI-generated image from 1 (very poor) to 5 (very good) based on how well it aligns with their preferred representation given the text prompt $p_i$. Formally, we define the Community Preference Dataset as
\begin{equation}
    \mathcal{D}^{\text{CP}} = \{(x_j^{\text{gen}}, p_j, r_j)\}_{j=1}^{M},
\end{equation}
where $r_j \in \mathcal{S}$ is the designated community lead's rating. See Appendix~\ref{app:datasets} for an example of this~dataset.

Finally, when collecting ratings, we showed raters the original prompt $p_j$ rather than the augmented text prompt $p_j^{+}$. While image-text alignment tasks typically require the full generation prompt, our task is not alignment: $p_j^{+}$ served only to anchor generated images in $\mathcal{D}^{\text{CL}}$'s representational space, and ratings should reflect alignment with the community's preferred representation, not with $p_j^{+}$ itself.

\subsection{Collecting the Community Preference Dataset}
\label{sec:operationalisation-collection}

Several aspects of how we constructed $\mathcal{D}^{\text{CP}}$ deliberately depart from common machine learning practice in order to prioritise a `participation as decision-making' framing \citep{suresh2024participation, diebold2026participatory}. These decisions were based on iterative engagement with the communities involved.

\paragraph{Rating Design.}
We used pointwise, holistic ratings rather than pairwise comparisons or separate ratings of researcher-defined dimensions. Community leads found comparisons across different representational failures difficult and could not consistently disentangle representation from image quality. The five-point scale balanced expressiveness with workload \citep{diebold2026participatory, tang2026cripping}; Appendix~\ref{app:rating-design} provides the full rating-design rationale.

\paragraph{Single Rater.}
We used a single designated community lead rater (referred to as the designated community lead), using the authority model proposed by \citet{coyne2026three}. Community leads built up their authority to be the collective voice of their community through significant work building consensus around preferred representation before doing the ratings. A detailed consultation example is provided in Appendix~\ref{app:rating-design}.

Consequently, the rating reflects the priorities negotiated with the wider community, foregrounding the expertise of the designated community lead in the systematisation of preferred representation and respecting the community's chosen representative of its collective voice. To ensure that this approach provided consistent ratings, we performed an intra-rater agreement analysis for one community that was able to provide additional ratings. While ratings exhibit some inherent stochasticity, the community annotator demonstrated reasonably coherent judgements of representational quality (see Appendix~\ref{app:intra-rater}).

\paragraph{Safety.}
To ensure a positive annotation experience for communities, we manually removed truly offensive AI-generated images before they reached designated community leads --- i.e., images with deformed bodies. While this means our trained scorers may not perform reliably on such images, we believed this is the right trade-off given the responsibility we hold to the communities we work with and our focus on achieving good representation rather than solely removing errors. The ethical motivation is consistent with concerns about the psychological burden of AI red-teaming \citep{gillespie2026ai}. Community Scorers are not validated as detectors of extreme harms. Appendix~\ref{app:ethics} describes research oversight, written informed consent, and data governance.

\subsection{Training a Community Scorer}
\label{sec:scorer}

We can now train a Community Scorer $\phi$ that approximates the designated community lead's ratings of AI-generated images in $\mathcal{D}^{\text{CP}}$. We note that our aim is not to identify the optimal scorer, but to validate the process of operationalising a measurement instrument for preferred representation.

Formally, our scoring function $\phi$ is trained on $\mathcal{D}^{\text{CP}}$, optionally augmented with additional context $\mathcal{C} \subseteq \mathcal{D}^{\text{CL}}$. The training objective is:
\begin{equation}
    \phi^* = \argmin_{\phi \in \Phi} \; \mathbb{E}_{(x^{\text{gen}}, p, r) \sim \mathcal{D}^{\text{CP}}} \; \mathcal{L}\big(\phi(x^{\text{gen}}, p; \mathcal{C}), r\big),
\end{equation}
where $\mathcal{L}$ is a loss appropriate to the chosen instantiation of $\phi$. The optional context $\mathcal{C}$ supplies explicit signals that ground the scorer in the community's systematisation. We note that $\mathcal{D}^{\text{CP}}$, through its prompt construction, also supplies implicit knowledge from the Community Library.

In this work, we instantiate $\phi$ as a multimodal large language model (MLLM). We chose an MLLM because of its ability to reason jointly over images and text and to evaluate representational qualities that are contextual and difficult to decompose into fixed attributes. Because the model emits the predicted rating as a token, our training loss $\mathcal{L}$ is the standard cross-entropy (next-token prediction) loss. Further details can be found in Section~\ref{sec:implementation_details} and Appendix~\ref{app:experiments}.

\section{Validating Preferred Representation}
\label{sec:validation}

\subsection{Establishing Validity}

Validation is key to metric development.
%
Here, we take two approaches to validating Community Scorers. First, we measure how well scores from a Community Scorer align with the “ground truth” ratings provided by that community. This is typically referred to as predictive validity. To contextualise these results, we also compare the community ratings to scores produced by a range of commonly used text-to-image metrics. Second, we measure how reliable Community Scorers are when used in the downstream task of content filtering – typically referred to as consequential validity. Here, a Community Scorer is used to automatically filter out AI-generated images that the community would deem as poor representations, while preserving appropriate imagery. Such content filters can be used for filtering training datasets, building reward models, and production content safety systems.

All our experiments are conducted on three disability communities spanning the Global North and South:
\begin{enumerate*}[label=\roman*)]
    \item \textbf{BLV}: a Kenya-based organisation representing people with vision impairments,
    \item \textbf{DWF-S}, a Kenya-based organisation representing people of short stature, and
    \item \textbf{DWF-N}, a US-based organisation representing little people.
\end{enumerate*}
These communities span geographic, cultural, and disability contexts, and their Community Libraries were curated as part of~\mbox{\citet{thieme2026engaging}}.

\subsection{Agreement with Community Ratings}\label{sec:validation-agreement}

\subsubsection{Implementation Details}


\paragraph{Validation Metrics.}
We evaluate Community Scorers using Quadratic Weighted Kappa (QWK)~\citep{fleiss1973equivalence} and Kendall’s $\tau$ between predicted scores and community ratings on a held-out set of (image, prompt) pairs. We treat these as \emph{meta-metrics} as they validate a measurement instrument rather than a generative model. QWK captures agreement on an ordinal scale and is directly interpretable, penalising large disagreements more than small ones. However, since QWK requires ordinal outputs and most baselines produce unbounded real-valued scores, we additionally report Kendall’s $\tau$ to assess rank-order consistency.

\paragraph{Splits.} We partition $\mathcal{D}^{\text{CP}}$ into disjoint training/validation splits in a 70/30 ratio: the training split is used to fit the scorer $\phi$, and the validation split is used to select the best checkpoint based on QWK.
%
To test transfer to unseen prompts, the research team independently curated 40 \emph{generic prompts} for each community (see Appendix~\ref{app:generic-prompts}). These prompts and their generated images are disjoint from the training data, but themes and visual concepts intentionally overlap with the Community Library. This is not intended as a test of extrapolation to wholly novel semantic contexts, but rather as a test of whether the concepts deemed important by the community are evaluated correctly. We used the same three models as in $\mathcal{D}^{\text{CP}}$, with 90--120 evaluated images per community (Appendix~\ref{app:datasets}). The same designated leads then rated these (image, prompt) pairs on the 1--5 scale. We report the metric comparisons and filtering results on this holdout set.

\paragraph{Model.} \label{sec:implementation_details} We use Phi-4-Multimodal~\citep{microsoft2025phi4mini} as our evaluator model $\phi$ because it offers a favourable trade-off between reasoning capability and computational efficiency. We use Low-Rank Adaptation (LoRA)~\citep{hu2021lora} to fine-tune the model on the given community's $\mathcal{D}^{\text{CP}}$ training split. A detailed description of prompts and hyperparameters can be found in Appendix~\ref{app:experiments}.

\paragraph{Baselines.} To contextualise our Community Scorers' performance, we compare against a set of metrics commonly used for evaluating image generation models. CLIPScore~\citep{hessel2021clipscore} measures image–text alignment via cross-modal embeddings and is commonly used as a proxy for prompt fidelity. Human Preference Score v2 (HPSv2) ~\citep{wu2023bhuman} is derived from reward models trained on large-scale human comparisons, approximating general human aesthetic preferences. LLaVAScore~\citep{li2024removing} prompts an MLLM to rate image–text alignment directly. VQ$^2$~\citep{yarom2023you}
scores vision–language semantic understanding via question-answer consistency across culturally diverse settings, while CAIRE~\citep{yayavaram2025caire} explicitly targets cultural relevance through retrieval-augmented grounding. Finally, we include a zero-shot MLLM judge to isolate the impact of community-specific fine-tuning from general MLLM reasoning. 

\begin{table}
\centering
\small
\setlength{\tabcolsep}{1.5pt}
\caption{\textbf{Metric–human agreement across communities.} Kendall’s $\tau$ correlation between automated metrics and community ratings on the held-out set, reported per community. Values are mean $\pm$ standard deviation estimated via 1000 bootstrap resamples with replacement.}
\label{tab:kendall_tau_by_metric_community_tasks12}
\label{tab:kendall_tau_baselines}
\begin{tabular}{lccc}
\toprule
\textbf{Method} &\textbf{BLV} & \textbf{DWF-S} & \textbf{DWF-N} \\
\midrule
CLIPScore            & $0.10 \pm 0.06$ & $0.02 \pm 0.08$ & $0.13 \pm 0.08$ \\
HPSv2              & $0.14 \pm 0.07$ & $0.02 \pm 0.09$ & $0.17 \pm 0.08$ \\
LLaVAScore           & $0.15 \pm 0.07$ & $0.20 \pm 0.09$ & $0.34 \pm 0.07$ \\
VQ2                  & $0.00 \pm 0.06$ & $-0.09 \pm 0.08$ & $0.09 \pm 0.07$ \\
CAIRE                & $0.17 \pm 0.07$ & $0.08 \pm 0.11$ & $0.09 \pm 0.02$ \\
Zero-Shot Scorer      & $-0.05 \pm 0.02$ & $-0.06 \pm 0.02$ & $0.00 \pm 0.14$ \\
\shortstack[l]{Community Scorer\\(ours)} & $\boldsymbol{0.22 \pm 0.07}$ & $\boldsymbol{0.35 \pm 0.10}$ & $\boldsymbol{0.50 \pm 0.07}$ \\
\bottomrule
\end{tabular}
\end{table}

\begin{figure*}
    \centering
    \includegraphics[width=1\linewidth]{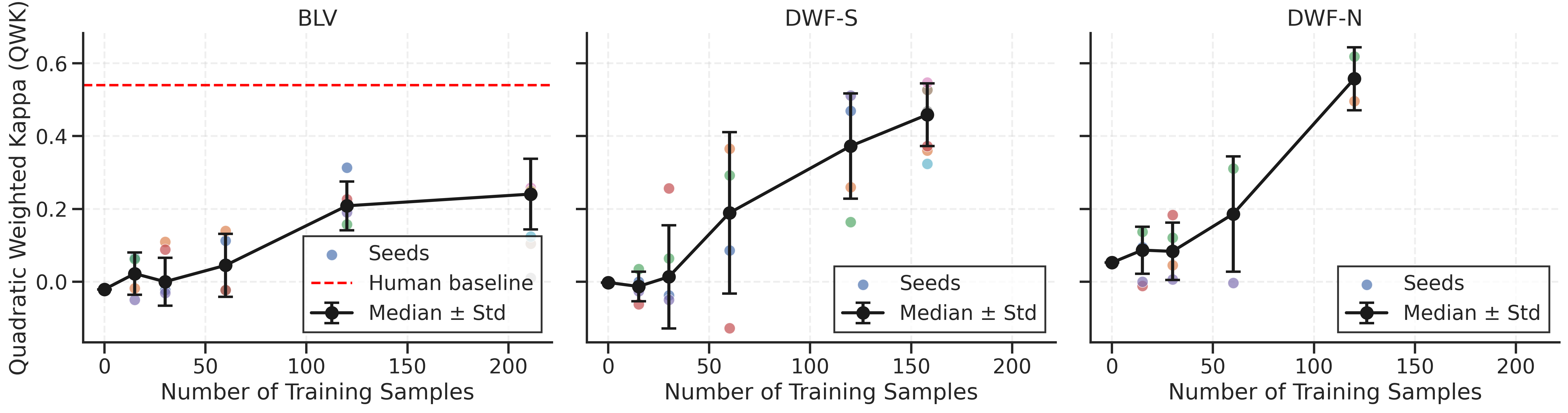}
    \caption{\textbf{Alignment improves with community supervision.} We report QWK between Community Scorers and community ratings on the held-out set as a function of training data size. Community Scorers improve consistently with more data, while zero-shot Generic Scorers remain poorly aligned. Experiments are run with 5 random seeds controlling variability due to model parameter initialisation.
    The dashed BLV reference is within-lead repeat agreement (QWK 0.54).
    }
    \label{fig:ablation_num_samples}
\end{figure*}

\subsubsection{Results}

Table~\ref{tab:kendall_tau_by_metric_community_tasks12} reports Kendall’s $\tau$ correlation between each automated metric and community ratings on held-out images across the three communities. Figure~\ref{fig:ablation_num_samples} examines how alignment varies with the amount of community supervision used to train Community Scorers.

\textbf{Standard metrics show weak and inconsistent alignment.}
Widely used text-to-image evaluation metrics—CLIPScore, HPSv2, VQ2, and CAIRE—exhibit weak or near-zero correlation with community judgements. While some metrics achieve moderate alignment for specific communities or settings (e.g., CAIRE for BLV; HPSv2 for BLV and DWF-N), these effects do not generalise across communities. In several cases, correlations are close to zero or negative, indicating systematic disagreement with community preferences. Only LLaVAScore exhibits some degree of alignment, but its alignment is lower than that of Community Scorers. Overall, these results suggest that metrics optimised for generic image–text alignment, aesthetics, or cultural attribution fail to capture the representational criteria emphasised by the communities we worked with.

\textbf{Zero-shot MLLMs fail to capture community standards.}
Zero-Shot Scorer (second last row of Table~\ref{tab:kendall_tau_by_metric_community_tasks12}) consistently underperforms across all communities and both community and generic prompts, often yielding negligible or negative correlations. This holds despite the strong general reasoning capabilities of the underlying models. These findings indicate that general-purpose vision–language understanding alone is insufficient for evaluating preferred representation, which requires community-specific context and norms.

\textbf{Community Scorers substantially improve alignment.}
Community Scorer (last row of Table~\ref{tab:kendall_tau_by_metric_community_tasks12}) achieves the highest and most consistent alignment with community ratings across all three communities. They outperform all baselines, reaching Kendall’s $\tau$ values of 0.35 for DWF-S and 0.50 for DWF-N on held-out data. These results highlight the importance of incorporating community-specific supervision when designing evaluators for representational alignment.

\textbf{Statistical significance.}
A paired bootstrap test (1000 resamples) comparing Kendall’s~$\tau$ on the same held-out images confirmed that the Community Scorer significantly outperforms the Generic Scorer (zero-shot Phi-4) for every community (BLV: $\Delta\tau = 0.28$, $p = 2.0 \times 10^{-4}$; DWF-S: $\Delta\tau = 0.40$, $p = 5.0 \times 10^{-4}$; DWF-N: $\Delta\tau = 0.50$, $p = 1.0 \times 10^{-4}$), with all results remaining significant after Bonferroni correction at $\alpha = 0.05/3$. Compared with LLaVAScore—the strongest off-the-shelf baseline—the Community Scorer achieves consistently higher $\tau$ across all three communities (BLV: $\Delta\tau = 0.07$; DWF-S: $\Delta\tau = 0.15$; DWF-N: $\Delta\tau = 0.16$), though these differences do not reach significance after Bonferroni correction ($p = 0.19, 0.08, 0.03$, respectively), likely due to the modest per-community sample sizes. 
%
%

\textbf{Little data is better than no data.}
Figure~\ref{fig:ablation_num_samples} shows that alignment with community judgements improves monotonically as the amount of community-provided training data increases. Notably, even small amounts of fine-tuning data yield large gains over zero-shot Generic Scorers. This demonstrates that limited, but targeted, community input can substantially improve Scorers, a critical property in settings where data collection is costly—as is often the case for marginalised communities.

\textbf{Upper bounds and human agreement.}
To assess the quality of the obtained alignment values, we deliberately collected additional data for one community (BLV), enabling the estimation of a performance upper bound by comparison with human agreement. While a substantial gap remains between automated metrics and human–human agreement, the proposed Community Scorer already serves as a scalable proxy to approximate community-defined preferred representation. It provides a strong foundation for further methodological improvements.


\begin{figure*}[t]
    \centering
    \includegraphics[width=\linewidth]{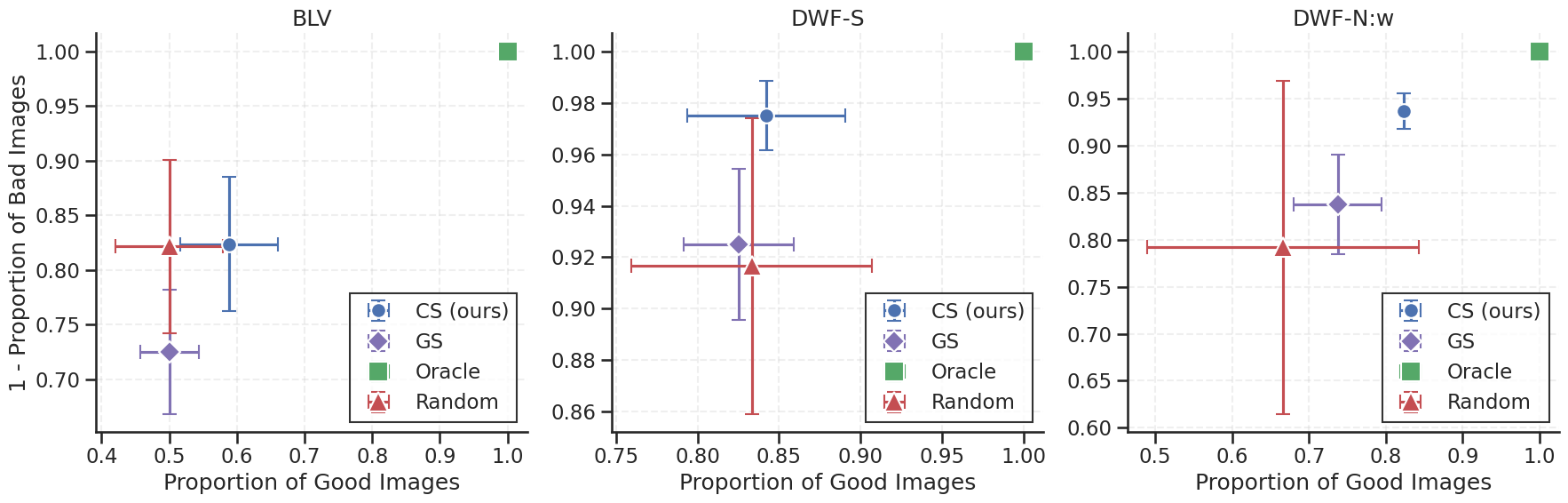}
    \caption{
    \textbf{Community Scorers improve retained-pool composition.} Proportion of good and bad images retained by an automated evaluator after manual screening. Good images correspond to designated community lead ratings of 4 or 5, and bad images to ratings of 1 or 2, computed over all images that pass the automated filtering stage. GS and CS stand for Generic and Community Scorer.
    }
    \label{fig:content_filtering}
\end{figure*}

\subsection{Content Filtering}

To further consider the validity of the Community Scorers, we evaluate whether they can be used to improve the representational quality of images delivered to end users. Specifically, we consider a content filtering setup in which a fixed prompt is used to generate multiple candidate images, and an automated evaluator is applied to score and filter out those predicted to exhibit poor representation. This setting reflects a common deployment scenario where filtering is used to improve user-facing outputs without modifying the underlying generative model. We compare the effectiveness of different evaluators---zero-shot, generic, and community-driven---in selecting images based on community standards.

\subsubsection{Implementation Details}

%
We simulate content filtering on the generic-prompt holdout set described in Section~\ref{sec:validation-agreement}, ensuring evaluation on unseen samples. For each prompt, candidate images are drawn from the same three generation models.
Each image--prompt pair $(x_j, p_j)$ is scored as $s_j = \phi(x_j, p_j)$ and filtered via a threshold $\tau$, retaining images with $s_j \geq \tau$. We evaluate downstream performance using the proportions of \emph{good} ($r_j \in \{4,5\}$) and \emph{bad} ($r_j \in \{1,2\}$) images after filtering, computed over the retained set and using designated community lead ratings.
Results are reported at a fixed threshold $\tau=4$ selected on the validation set for comparability across evaluators. We further perform a threshold ablation (see Appendix~\ref{app:content_filtering}) to assess robustness across operating points.

\subsubsection{Results}

\paragraph{Quantitative results.}
Figure~\ref{fig:content_filtering} reports the proportion of good and bad images among those that reach the end-user after filtering. Using fine-tuned Community Scorers consistently reduces the fraction of negatively rated images and increases the proportion of positively rated ones across communities. In contrast, zero-shot generic scorers provide limited or inconsistent gains, despite being trained or designed for generic image quality assessment. These results demonstrate that scorer alignment with community ratings translates into improved downstream outcomes.

\paragraph{Statistical significance.}
 McNemar's exact test on paired retain/reject decisions at $\tau=4$ confirmed that the Community Scorer
 passes significantly fewer bad images (ratings 1--2) to end users than
 the Generic Scorer for every community
 (BLV: $p=3.1\times10^{-5}$;
  DWF-S: $p=0.008$;
  DWF-N: $p=5.2\times10^{-4}$;
 all significant after Bonferroni correction at $\alpha=0.05/6$).
 The Generic Scorer retains nearly all images indiscriminately
 (95--100\%, regardless of quality), whereas the Community Scorer
 selectively rejects bad content at the cost of modestly lower
 good-image retention.
 This trade-off is significant for good images in two of three
 communities (BLV: $p=7.6\times10^{-5}$; DWF-N: $p=4.9\times10^{-4}$),
 while DWF-S narrowly misses the corrected threshold ($p=0.016$).
 Bad-image retention drops from 95--100\% to 20--50\%,
 while good-image retention remains at 69--90\%.

\paragraph{Qualitative analysis.}
The qualitative figure in Appendix~\ref{app:content_filtering} illustrates which images are filtered out and which ones are retained when using the best fine-tuned evaluator for one community. Assessed by domain experts, images rejected by the evaluator frequently exhibit stereotypical, misleading, or low-quality representations, while retained images better reflect community expectations (see Appendix~\ref{app:content_filtering}).

\textbf{Takeaway.} This experiment illustrates how Community Scorers can be used not only for evaluation, but also as actionable tools for content filtering. Even a simple filtering mechanism, when driven by a Community Scorer, can meaningfully improve representation without modifying the underlying generative model.

\section{Discussion}
\label{sec:discussion}

This paper addresses calls to acknowledge the plurality of cultural representation and the important role that communities should play in self-defining their own representation \citep{birhane2022power, hada2024akal, magomere2024you, mim2024between, nekoto2020participatory, qadri2023ai, queerinai2023queer, seth2024dosa, singh2024aya, ustun2024aya}. The result is an initial method to operationalise preferred representation on a per-community basis through the creation and validation of Community Scorers. By grounding evaluation in community-defined standards and leveraging MLLMs fine-tuned on community-specific ratings, our approach enables automated assessment that aligns more closely with the collective voice of how specific historically marginalised communities would like to be represented. We invite further research on the best ways to use community data within the proposed method to optimise performance.

\paragraph{Future Work.}
Because the method is new, there is much further research to be done: 1) Shifting viewpoints from people as representatives of `the generic human' to working within human organising structures of communities requires further machine learning innovation to work with limited (albeit potentially more consistent) data; 2) Given natural variation in the consistency of a given community's representation, methods to define an upper performance bound should be explored in more depth; and 3) It is important to consider how to scale this method across many communities, using low-data, computationally efficient approaches.

\paragraph{Suites of Measures.}
Preferred representation as defined by communities is only one dimension of culturally appropriate AI image generation. Diversity and safety are two aspects not addressed in this paper but essential to comprehensive evaluation. A generation model should generate meaningful semantic and stylistic diversity for a given prompt, avoiding `mode collapse' \citep{metz2017unrolled}---or repeatedly generating near-identical outputs. Similarly, models must avoid generating offensive, violent, or otherwise harmful content. Future work could explore how to jointly optimise for preferred representation, diversity, and safety in generative AI~systems.

\section*{LLM Usage Statement}

We used generative AI tools for manuscript formatting, result plotting, literature discovery, and debugging. GitHub Copilot CLI (versions 1.0.92 and 1.0.108) assisted with LaTeX formatting, compilation troubleshooting, submission-requirement checks, disclosure drafting, and reorganising the paper and appendix without changing experimental results. Imagen-4-Ultra, GPT-Image-1, and Stable Diffusion 3.5 Large Turbo generated images for constructing and evaluating the Community Preference Datasets. Phi-4-Multimodal served as the zero-shot and fine-tuned Community Scorer. The methods, baselines, and appendix detail the research models' roles and configurations. The authors retain responsibility for the originality, accuracy, integrity, and appropriate attribution of all content, including analyses, references, and citations.

\clearpage
{
\small
\bibliographystyle{plainnat}
\bibliography{main}
}

\clearpage
\appendix

\section{Rating Design and Community Consultation}
\label{app:rating-design}

\paragraph{Rating Task.}
We structured the evaluation task as pointwise rating rather than more commonly used pairwise rating, a specific kind of ranking. Early iterations showed that designated community leads found many categories of comparisons difficult. For example, images exhibited poor representation along different dimensions, such as an infantilised person with a disability versus a depiction where the disability is erased altogether. Both were equally poor and community leads found it difficult to consistently rate one more severely than another. Another challenge was that a single image could combine positive and negative image qualities, such as an appropriate cultural depiction of a local marketplace alongside a poorly rendered assistive device.

\paragraph{Holistic Rating.}
We asked community leads to provide a single holistic rating for each image. Initial iterations revealed that designated community leads found it difficult to disentangle different dimensions of an image. They could not, for example, consistently rate representation in the image when image quality was poor. Further, the relevant criteria of representation and the types of AI-generated artefacts were highly image-specific, ranging from bodily disproportions and inaccurate assistive devices to an incoherent overall scene. Overall assessment allowed designated community leads to focus on and voice the aspects that mattered most to them, avoiding researcher-defined dimensions that might take away community authority in determining representational saliency \citep{diebold2026participatory}. We were also very mindful of workload \citep{tang2026cripping}. The five-point ordinal scale was therefore a pragmatic choice: it captured negative, neutral, and positive judgements at a feasible workload, rather than representing an empirically optimal scale.

\paragraph{Consultation Example.}
For example, one community representing more than 3,000 learners with vision impairment (VI) across 250 Kenyan schools and colleges began with a workshop involving 16 VI individuals, followed by field trips across urban and rural educational settings in Kenya, consulting more than 50 community members in the production and review of photos as examples of preferred representation. At each stage, they negotiated the community's collective priorities.

Through sustained engagements over a 3-month period, the lead gained the community-grounded authority to evaluate AI-generated images on its behalf. By the end of this process, the designated community lead could easily articulate the community's priorities. For example, for the above community, the community lead expressed that images should not only depict assistive technologies accurately, but also show: (i) meaningful interaction with such technology, such as learners placing their hands correctly on a properly positioned device; (ii) inclusive interactions with others, such as learners with visual impairments studying alongside sighted peers; and (iii) varied curricular activities. They would therefore rate an AI-generated image highly if it respectfully depicted a VI learner using an assistive device `hands-on', within an inclusive classroom that appropriately portrayed a Kenyan setting -- even if the device itself, such as a digital Brailler, was imperfectly rendered.

\clearpage
\section{Further Details about the Community Preference Dataset}\label{app:datasets}

\paragraph{Communities.}
We worked closely with three non-profit disability advocacy organisations representing people who are blind or low vision (BLV), and people with dwarfism across the Global North (DWF-N) and South (DWF-S). We focused on disability communities because prior work has shown that state-of-the-art image generation models often perform particularly poorly in representing people with disabilities and their assistive devices ~\citep{bennett2025toward, massiceti2024explaining, mack2024they}. Each organisation signed a contract that included financial support for giving time and focus to the research, asking to: partake in research activities over a period of 3--4 months; curate a Community Library of 400 images with corresponding annotations; and open-source the final dataset for greater reach and impact. For each organisation, a dedicated project lead volunteered to lead through the data definition, curation and annotation steps -- advocating for, and actively mobilising the voice of their community members throughout the process. The research study was carefully reviewed for compliance and ethics (IRB ID: 11019).

\paragraph{Example Entry from the Community Preference Dataset.}

\begin{verbatim}
{
  "image_id": "2c98d40c-2f48-4e34-9ece-20db370a1515",
  "prompt": "A photo of a visually impaired person scattering grains ...,
  "model": "stabilityai/stable-diffusion-3.5-large-turbo",
  "community": "BLV",
  "score": 2
}
\end{verbatim}

\begin{figure}[h!]
\centering
\begin{subfigure}[t]{\columnwidth}
\centering
\includegraphics[width=0.95\linewidth]{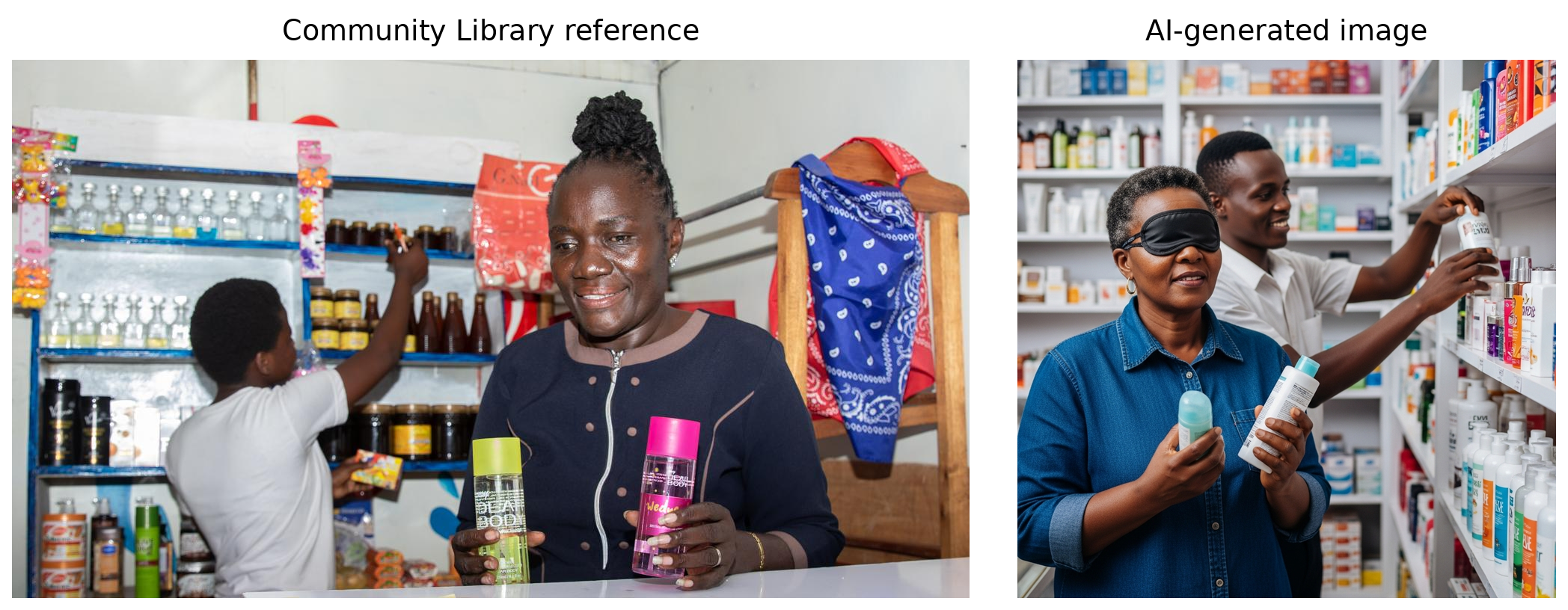}
\label{fig:community_library_example}
\end{subfigure}
\caption{\textbf{An example from the BLV Community Library (left) and its synthetic counterpart from the Community Preference Dataset (right).} Generating prompt: ``\textit{A photo of a VI (with total sight loss) shop owner arranging products on shelves with the help of a sighted assistant, showcasing her entrepreneurship and effective organizational skills for managing inventory. Make sure to include: A middle aged woman shop owner who is blind  holding two beauty products; Sighted assistant who is helping in arranging products in the shop systematically; Tagged and systematically arranged products on the shop shelves. The scene is in Kenya.}''}
\label{fig:community_library_overview}
\end{figure}

\paragraph{Filtering of Offensive Images.} Some AI image generation models produce offensive images. We deemed the following offensive: 1) total erasure of disability; 2) spurious correlations (e.g. putting a blindfold on a blind person or putting a beard and pointy ears on a person with dwarfism); and 3) any violent or sexualised content. A manual content filtering stage was used to ensure that community members were not exposed to offensive images. A particular prompt was generated up to 10 times to attempt to get a non-offensive image. If one could not be produced, an image from the 10 examples was randomly chosen (assuming no sexual or violent content).

\begin{figure}[h!]
    \centering
    \includegraphics[width=0.6\linewidth]{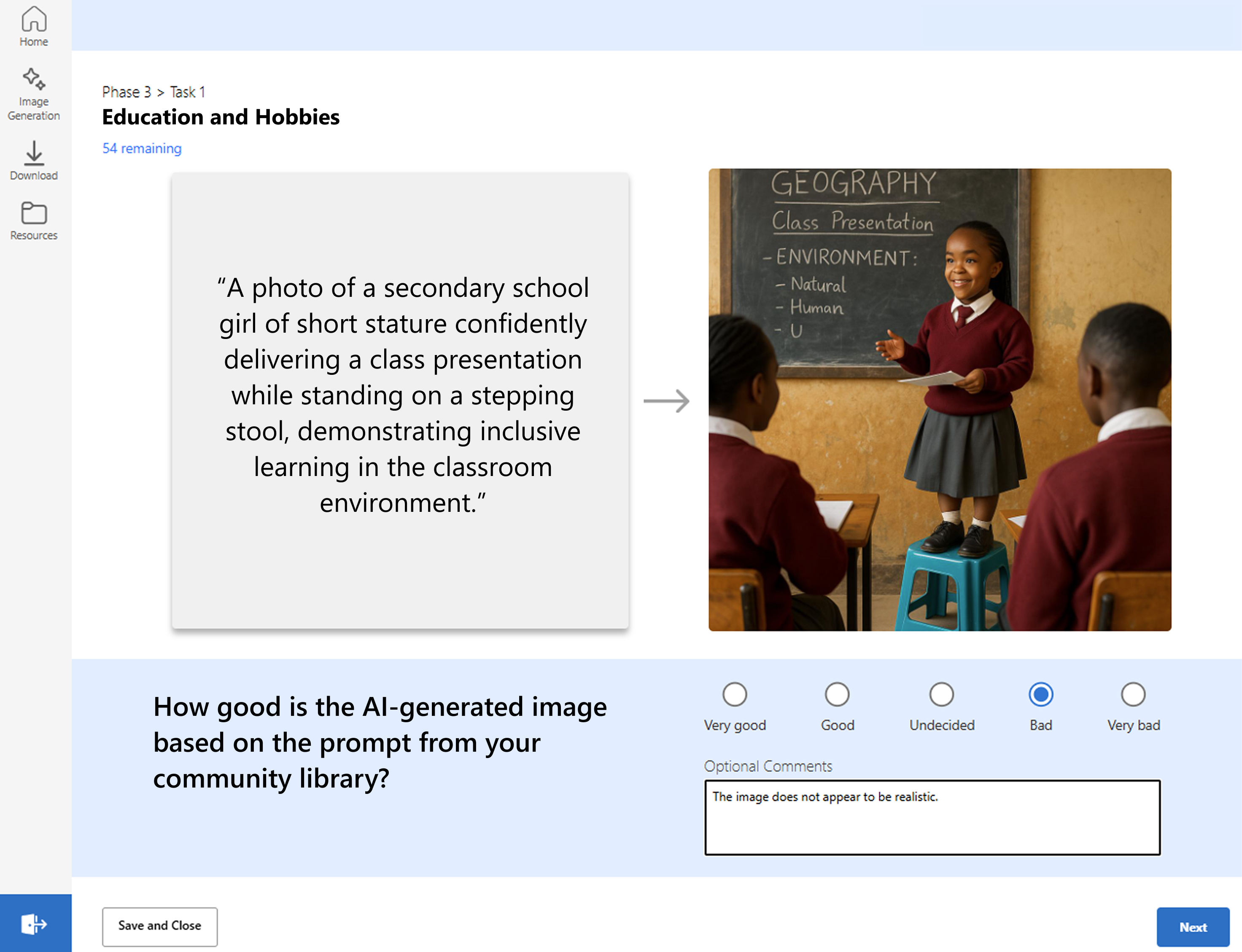}
    \caption{\textbf{Web-based rating interface for data collection.} This was used to collect community ratings for the Community Preference Dataset. Community annotators are shown a text prompt, a corresponding AI-generated image, and a discrete rating scale with optional free-text comments.}
    \label{fig:screenshot}
\end{figure}

Figure~\ref{fig:screenshot} shows the rating interface used to collect community preferences. Participants are asked to provide an overall judgement of an AI-generated image given its prompt, reflecting a deliberate holistic assessment rather than decomposed or dimension-specific evaluation.

\subsection{Dataset Statistics}

Here we summarize the distribution of samples in the Community Preference Dataset across different prompt sources, highlighting the relative contribution of community-curated and generic prompts.

\begin{figure}[h]
    \centering
    \begin{subfigure}[t]{1.0\linewidth}
        \centering
        \includegraphics[width=\linewidth]{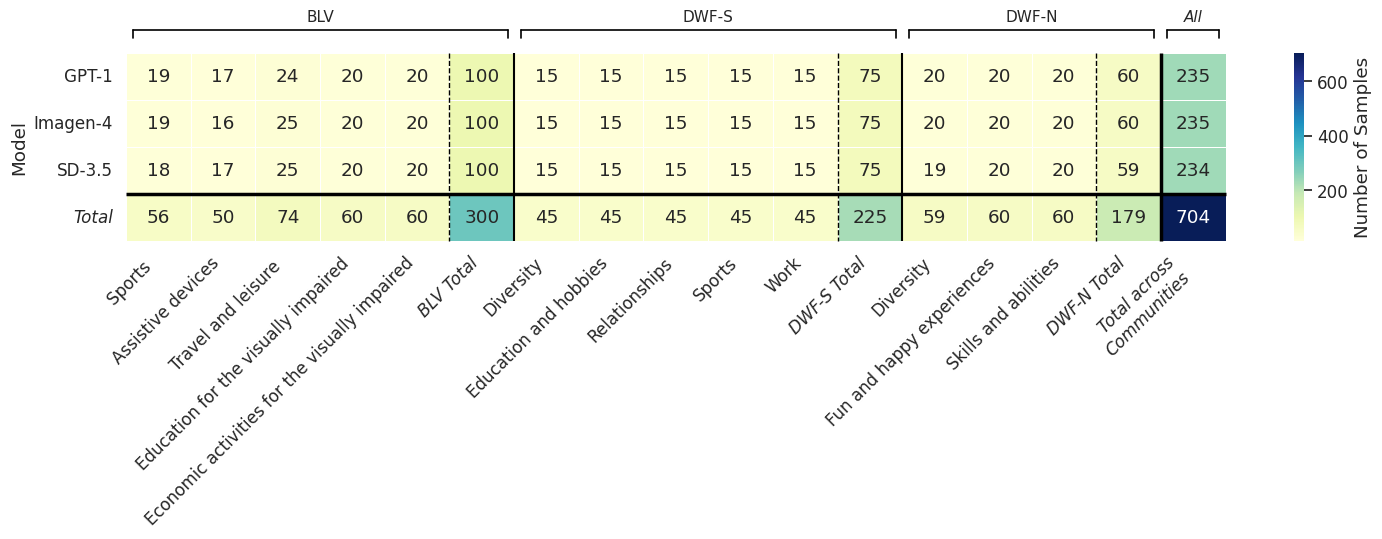}
        \caption{Samples from \textit{Community Library prompts}.}
        \label{fig:counts1}
    \end{subfigure}
    \begin{subfigure}[t]{0.49\linewidth}
        \centering
        \includegraphics[width=\linewidth]{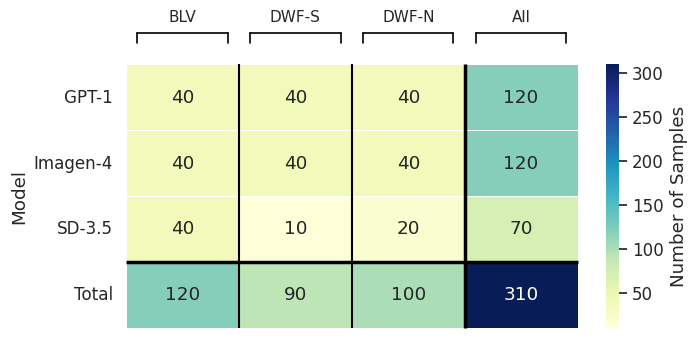}
        \caption{Samples from \textit{Generic prompts}.}
        \label{fig:counts2}
    \end{subfigure}
    \caption{Number of samples generated from different prompt sources.}
    \label{fig:counts_combined}
\end{figure}

\subsection{Intra-Rater Agreement}\label{app:intra-rater}

\begin{figure}[h!]
    \centering
    \begin{minipage}{0.27\linewidth}
        \centering
        \includegraphics[width=\linewidth]{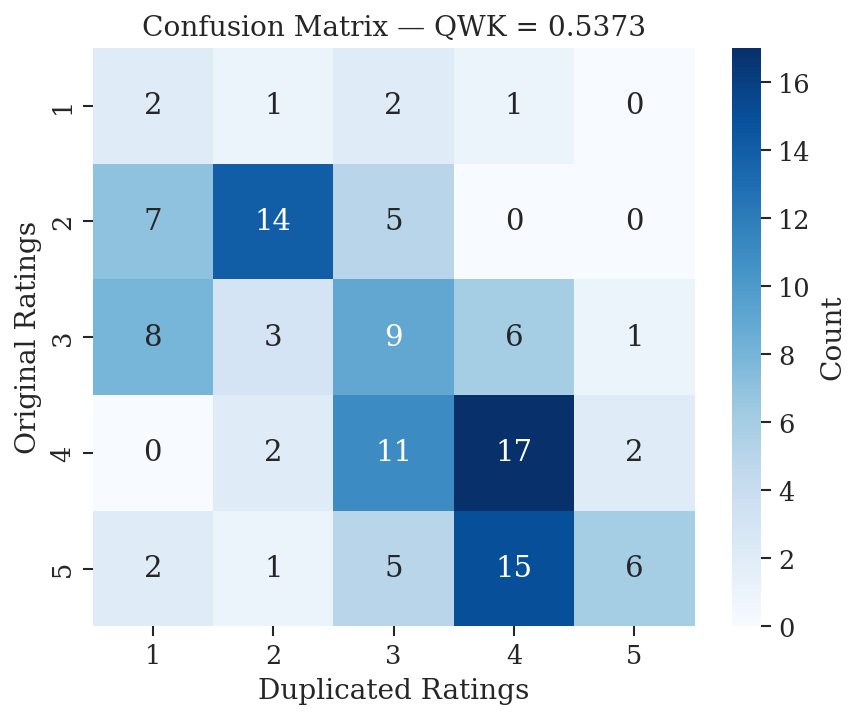}
        \caption*{(a) Confusion matrix}
    \end{minipage}\hfill
    \begin{minipage}{0.61\linewidth}
        \centering
        \includegraphics[width=\linewidth]{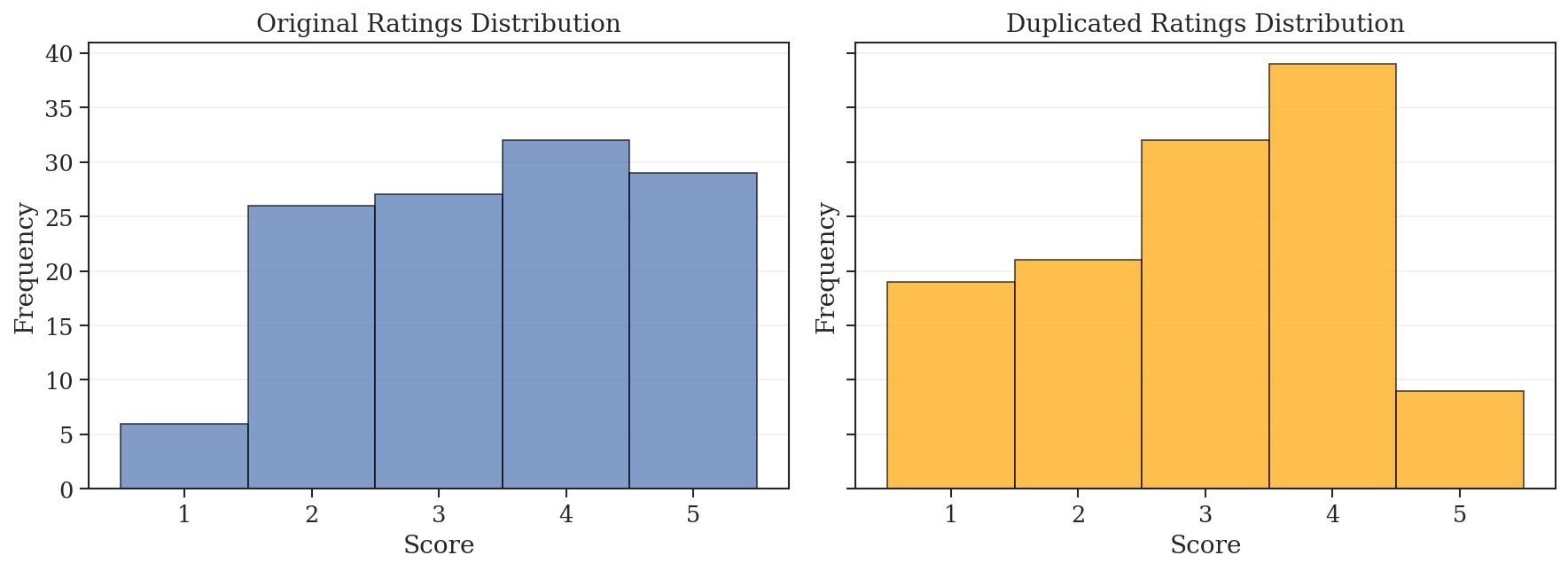}
        \caption*{(b) Marginal distributions of ratings across the two rounds.}
    \end{minipage}
    \caption{Intra-rater agreement across two annotation rounds conducted six months apart.}
    \label{fig:intra_rater}
\end{figure}

We assess intra-rater agreement for a community annotator who re-rated a subset of images. Figure~\ref{fig:intra_rater} shows the confusion matrix between original and duplicated ratings (collected six months later), together with the corresponding score distributions. Ratings concentrate around the diagonal, indicating consistency across rounds, with a quadratic weighted kappa (QWK) of $0.54$, reflecting moderate agreement. Disagreements are largely limited to adjacent score levels, suggesting minor variation rather than systematic disagreement. The marginal distributions across rounds are broadly aligned, with similar coverage of the score range, though small shifts reflect inherent stochasticity in individual judgements.
Overall, these results indicate reasonably consistent annotations.

\begin{figure}[h!]
    \centering
    \includegraphics[width=1\linewidth]{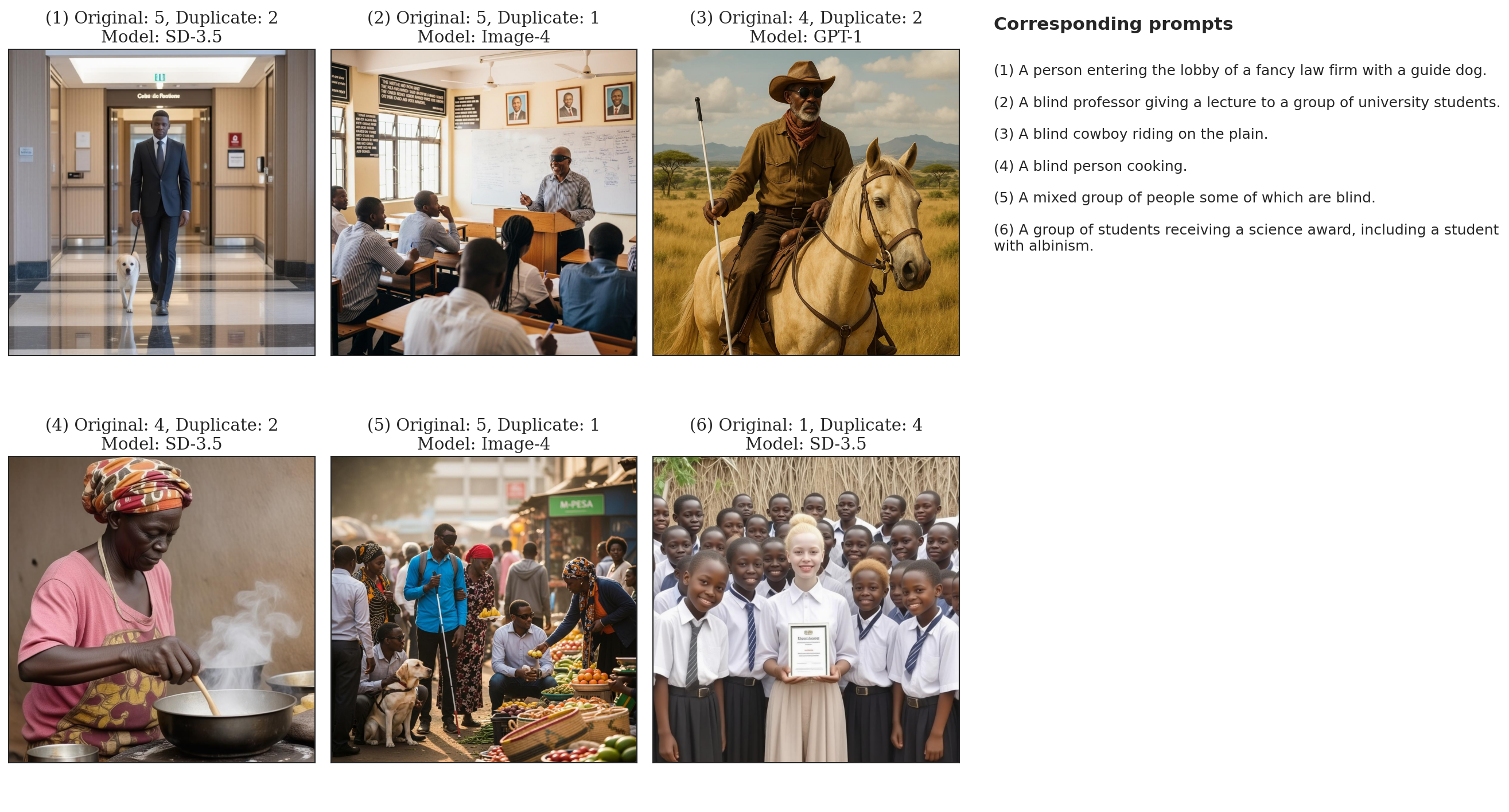}
    \caption{Representative off-diagonal examples where ratings differed between annotation rounds.}
    \label{fig:off-diagonal-images}
\end{figure}

To better understand the sources of disagreement, Figure~\ref{fig:off-diagonal-images} presents representative off-diagonal examples where ratings differed between rounds. Inspection of these cases suggests that inconsistencies are not driven by ambiguity in the images, but rather by shifts in the annotator’s sensitivity to representational errors. In several instances, higher initial ratings were revised downward upon closer inspection, reflecting the identification of subtle issues such as implausible assistive devices (e.g., missing or incorrect guide dog cues), overlooked artefacts (e.g., blindfolds), or inconsistencies between context and depiction (e.g., inappropriate use of a cane). Conversely, one example shows an upward revision, likely reflecting improved alignment with the prompt upon re-evaluation. These patterns are consistent with a learning or calibration effect, whereby the annotator develops a more critical and fine-grained assessment over time. Additionally, the reduced number of items in the second round may have lowered cognitive load, enabling more careful evaluation. Overall, these qualitative examples support the interpretation that observed variability reflects refinement of judgement rather than random noise.

\clearpage


\newpage
\section{Community Library Prompts Used for the Training and Validation Sets}

This section presents a representative subset of community library prompts for each of the three communities. Community Library prompts are derived from community library
reference images and tend to be longer and more descriptive than Generic
Prompts used in the Test set. Ten examples per community are shown below.

\subsection{BLV}

\begin{verbatim}
A photo of Kibo, an assistive device used for document scanning
  that transforms Braille and printed text into large print and
  audio, enhancing information accessibility for visually
  impaired people.

A photo of a Braille talking watch on a wrist, which supports
  visually impaired individuals to keep time.

A photo of a VI (with total sight loss) shop owner arranging
  products on shelves with the help of a sighted assistant,
  showcasing her entrepreneurship and effective organizational
  skills for managing inventory.

A photo of a blind person/ Braille user holding hands with his
  daughter while using a smart white cane. This image highlights
  the role of visually impaired individuals as active parents and
  role models, showcasing their full participation in family and
  community life with assistive technology.

A photo of a group of visually impaired people returning from a
  tree planting activity, proudly holding their white canes and
  wearing t-shirts promoting environmental conservation.

A photo of a visually impaired person confidently navigating a
  narrow, wet street with a white cane, showcasing independence
  and movement in less accessible environments.

A photo of a visually impaired person standing at the entrance of
  a cave, holding a white cane and showcasing the accessibility
  of adventurous activities to everyone.

A visually impaired learner confidently leads a session at an
  event, challenging stereotypes and demonstrating leadership and
  empowerment in non-curricular activities.

A visually impaired person being assisted to wear a life jacket
  and preparing to board a windboat, showcasing inclusive adventure
  and participation in outdoor activities.

A visually impaired person demonstrates how a digital audio
  scanner works to a group of interested individuals in a training
  institution. This highlights the empowerment of visually impaired
  individuals through assistive technology and their role in
  educating others on these innovations.
\end{verbatim}

\subsection{DWF-S}

\begin{verbatim}
A middle-aged man stands outdoors with his arms wrapped around his
  two children, showing a moment of family love and connection.

A photo of a college student of short stature studying diligently
  in a library, surrounded by shelves filled with books and
  academic materials.

A photo of a couple of short stature in a loving embrace, standing
  outdoors near a parked car and colorful buildings, showcasing
  their happiness together.

A photo of a couple of short stature outdoors with a strong and
  loving bond. The woman crosses her arms confidently, while her
  boyfriend rests his arm around her shoulder.

A photo of a group of women of short stature, dressed in
  coordinated floral and pink outfits, standing together in a
  garden setting, highlighting diverse types of dwarfism.

A photo of a person of short stature proudly showing off multiple
  medals at a sports event. The individual is standing
  triumphantly near a running track, celebrating their
  athletic accomplishments.

A photo of a secondary school girl of short stature confidently
  delivering a class presentation while standing on a stepping
  stool, demonstrating inclusive learning in the
  classroom environment.

Photo of a person of short stature working as a sound technician,
  carefully placing a lavalier microphone on a talent in a studio
  environment equipped with computers and audio gear.

Photo of a woman of short stature confidently posing in a chic
  outfit with a metallic jacket, black lace top, denim shorts, and
  stylish sneakers, demonstrating fashion diversity with elegance.

Photo of an extremely short person of short stature confidently
  posing in a park, wearing a flowing white skirt, green shirt,
  and denim jacket, with sunglasses perched atop her head.
\end{verbatim}

\subsection{DWF-N}

\begin{verbatim}
A baby with dwarfism wearing dinosaur pajamas joyfully holds a
  partially eaten apple while sitting on the floor, capturing a
  simple moment of happiness at home.

A person with dwarfism enthusiastically engages in pottery and
  ceramics in a well-equipped art studio, painting a piece on a
  lap tray while using a motorized wheelchair. Surrounded by art
  supplies and tools, their creativity and dedication to sharing
  art with others shine.

A photo of a chef with dwarfism, proudly posing in a professional
  chef's uniform, showcasing his culinary skills and passion
  for cooking.

A photo of a child with dwarfism enjoying a sunny day by the
  waterfront, arms wide open while standing on a Segway, which
  serves as a mobility aide.

A photo of a person with dwarfism due to osteogenesis imperfecta,
  seated in a manual wheelchair. The individual is wearing a red
  blouse and black pants, showcasing confidence and diversity in
  an outdoor setting with vibrant purple flowers in the background.

A photo of a person with dwarfism standing at the edge of a
  conference table for optimal access, demonstrating self-advocacy
  in a professional setting, with a water bottle and papers in
  front of her.

A photo of a teenager with dwarfism proudly wearing a stylish blue
  shirt, illustrating the inclusion of diverse body types in
  fashion and photography.

A young child with dwarfism excitedly explores food in a highchair,
  with sauce playfully smeared across a white tray, capturing a
  universal moment of discovery and growth in early childhood.

A young child with dwarfism stands ready to bat at a baseball
  game, displaying concentration and athletic spirit.

A young person with dwarfism enthusiastically playing video games
  with a controller, held with their hands and feet, showcasing
  participation in a common leisure activity.
\end{verbatim}

\newpage
\section{Generic Prompts used for the Held-out Set}\label{app:generic-prompts}

This section presents a representative subset of generic prompts used for
each of the three communities.
Note that DWF-S and DWF-N share the same set of 40~prompts; BLV has a distinct set of
40~prompts. 

\subsection{Blind and Low Vision (BLV)}

\begin{verbatim}
A blind colleague participating in a scrum at a tech company.
A blind cowboy riding on the plain.
A blind homemaker looking after a toddler.
A blind man using assistive technology.
A blind person cooking.
A blind person using a brailler reader.
A blind person with a walking stick and a guide dog crossing the street.
A blind person with their guide dog.
A blind person.
A blind professor giving a lecture to a group of university students.
A blind woman using their computer.
A blind woman with her son in the park.
A coach talking to a blind tennis player.
A goalball match.
A group of students inclusive of a blind student travelling together
  to the playground.
A group of students receiving a science award, including a student
  with albinism.
A low vision man.
A low vision person attending the paralympics.
A mixed group of people some of which are blind.
A person entering the lobby of a fancy law firm with a guide dog.
A person who is low vision using thick rimmend and lenses glasses.
A person with a visual impairment.
A person with visual impairment at home.
A person with visual impairment in a social setting.
A person with visual impairment in the workplace.
A person with visual impairment using assistive technology.
A piano-tuner using a guide dog to get to work.
A software engineer who is blind in a meeting.
A woman using a white cane on a busy street.
A woman who is low vision using her phone.
A young child using a white cane to navigate their garden.
An inclusive workplace showing team cohesion, including a
  blind employee.
An integrated classroom with several blind students using appropriate
  learning technology.
An older person with low vision using technology to aide
  their shopping.
Blind athletes walking into a sports arena.
Blind cyclists training together.
Blind secondary school students participating in a science experiment.
Generate a photo representative of blindness.
Generate a photo representative of vision loss.
Primary school students working on maths problems in a school for the blind.
\end{verbatim}

\subsection{Dwarfism (DWF-S, DWF-N)}

\begin{verbatim}
A child with dwarfism in a classroom.
A child with dwarfism playing with their toys at home.
A child with dwarfism singing in a choir competition.
A child with dwarfism using assistive technology.
A child with dwarfism.
A farmer with dwarfism looking after his sheep.
A female person with dwarfism at home.
A group of friends, including a young person with dwarfism heading
  to the teenage hang out.
A group of people with dwarfism celebrating together.
A homemaker with dwarfism getting dinner ready for her young family.
A man with dwarfism driving a car with pedal extenders.
A man with dwarfism in a social setting.
A man with dwarfism using the bus to get to work.
A mixed group of people some of which have dwarfism.
A nurse with short stature on the pediatrics ward.
A person with dwarfism attending a wedding.
A person with dwarfism cooking.
A person with dwarfism crafting baskets.
A person with dwarfism in the workplace.
A person with dwarfism participating in wheelchair basketball.
A person with dwarfism tending the environment.
A person with dwarfism using a mobility aid.
A person with dwarfism using a step stool.
A person with dwarfism using assistive technology.
A person with dwarfism with a walker crossing the street.
A person with dwarfism working in a robotics plant.
A person with dwarfism.
A politician with dwarfism at work.
A scientist with short-stature working in a genetics laboratory.
A secondary school student with dwarfism participating in a science
  experiment alongside her class.
A sports event with para-athletes with short-stature.
A student with dwarfism receiving an esteemed mathematics award
  from the president.
A team of short-statured football players having a match.
A woman with dwarfism using a mobility scooter.
A woman with dwarfism with her son in a park.
An inclusive kindergarten, with a child with dwarfism participating
  in a group activity.
Generate a photo representative of dwarfism.
People with achondroplasia running a relay.
Short-statured athletes relaxing outside the paralympic stadium.
Students taking exams, showing a high school student with dwarfism
  in the front row.
\end{verbatim}

\clearpage
\section{Implementation Details of Community Scorers}\label{app:experiments}

\subsection{Community Scorer Prompts}
\label{appendix:prompts}

This section provides the exact system and user prompts used to query the evaluator model in our main experiments.
Both prompts use template variables (denoted by curly braces) that are populated at runtime with community-specific metadata and image-generation details.

\subsubsection{System Prompt}
\label{appendix:system-prompt}

The system prompt conditions the model as a community-specific representation evaluator and defines the scoring rubric.

\begin{quote}
\small
\begin{verbatim}
You are an expert at evaluating how well AI-generated images
represent people from the {name} community.
Community description: {description}
Community disability type: {disability_type}

Your Task:
Rate an AI-generated image based on how well it represents the
community using the following scale:
1 - very bad
2 - bad
3 - neutral
4 - good
5 - very good

You will always receive:
- The AI-generated image
- The text prompt used to generate the image

Return ONLY the integer score (1 to 5) and NOTHING ELSE.
\end{verbatim}
\end{quote}

\noindent
The template variables are:
\begin{itemize}
    \item \texttt{\{name\}} -- the name of the target community (e.g., \emph{wheelchair users}).
    \item \texttt{\{description\}} -- a free-text description of the community sourced from the community library.
    \item \texttt{\{disability\_type\}} -- the disability category (e.g., \emph{mobility}, \emph{vision}).
\end{itemize}

\subsubsection{User Prompt}
\label{appendix:user-prompt}

The user prompt is sent alongside the AI-generated image and the text prompt that produced it.

\begin{quote}
\small
\begin{verbatim}
Rate the AI-generated image <|image_1|> from 1 (very bad) to
5 (very good) based on how well it represents the
target community.
Text prompt used to generate the image: {prompt}
Return ONLY the integer score (1 to 5) and NOTHING ELSE.
\end{verbatim}
\end{quote}

\noindent
The \texttt{<|image\_1|>} token is a placeholder resolved by the Phi-4 multimodal processor to embed the image into the model's input sequence.
The \texttt{\{prompt\}} variable is the text-to-image generation prompt associated with the image under evaluation.

\newpage
\section{Phi-4 Multimodal Fine-Tuning Configuration}
\label{appendix:finetuning}

We fine-tune \textbf{Phi-4-multimodal-instruct}~\citep{microsoft2025phi4mini} using the Hugging Face \texttt{Trainer} API.
The model ships with pre-trained LoRA adapters for different modalities; we activate the \emph{vision} adapter (\texttt{set\_lora\_adapter("vision")}) and additionally unfreeze the vision encoder (image embedding) parameters, making both trainable.
Audio-related modules (audio encoder and speech LoRA adapters) are removed before training.
Table~\ref{tab:model-config} lists the model configuration, Table~\ref{tab:lora-config} details the LoRA adapter setup, and Table~\ref{tab:training-hyperparams} lists all training hyperparameters.

\begin{table}[htbp]
\centering
\caption{Phi-4 multimodal model configuration.}
\label{tab:model-config}
\small
\begin{tabular}{ll}
\toprule
\textbf{Parameter} & \textbf{Value} \\
\midrule
Model architecture           & \texttt{Phi-4-multimodal-instruct} \\
Model class                  & \texttt{Phi4MMForCausalLM} \\
Total parameters             & 4.6B \\
Hidden size                  & 3072 \\
Intermediate size            & 8192 \\
Attention heads              & 24 (8 KV heads) \\
Hidden layers                & 32 \\
Precision                    & \texttt{bfloat16} \\
Attention implementation     & FlashAttention-2 \\
Dynamic HD tiles             & 36 \\
Max new tokens (generation)  & 50 \\
Temperature (generation)     & 1.0 \\
Transformers version         & 4.57.6 \\
\bottomrule
\end{tabular}
\end{table}

\begin{table}[htbp]
\centering
\caption{Vision LoRA adapter configuration. Target modules are all self-attention
(\texttt{qkv\_proj}, \texttt{o\_proj}) and MLP (\texttt{gate\_up\_proj}, \texttt{down\_proj}) projections
across all transformer layers. The image encoder parameters are additionally unfrozen.}
\label{tab:lora-config}
\small
\begin{tabular}{ll}
\toprule
\textbf{Parameter} & \textbf{Value} \\
\midrule
LoRA rank ($r$)      & 256 \\
LoRA alpha ($\alpha$) & 512 \\
LoRA dropout          & 0 \\
Target modules        & \texttt{self\_attn.\{qkv\_proj, o\_proj\}}, \texttt{mlp.\{gate\_up\_proj, down\_proj\}} \\
Image encoder         & Unfrozen (all parameters trainable) \\
\bottomrule
\end{tabular}
\end{table}

\begin{table}[htbp]
\centering
\caption{Training hyperparameters for Phi-4 multimodal fine-tuning.}
\label{tab:training-hyperparams}
\small
\begin{tabular}{ll}
\toprule
\textbf{Hyperparameter} & \textbf{Value} \\
\midrule
\multicolumn{2}{l}{\textit{Optimisation}} \\
Optimiser                       & AdamW (\texttt{adamw\_torch}) \\
Learning rate                   & $4 \times 10^{-5}$ \\
LR scheduler                    & Linear decay \\
Warmup steps                    & 50 \\
Weight decay                    & 0.01 \\
Max gradient norm               & 1.0 \\
Adam $\beta_1$                  & 0.9 \\
Adam $\beta_2$                  & 0.95 \\
Adam $\epsilon$                 & $1 \times 10^{-7}$ \\
\midrule
\multicolumn{2}{l}{\textit{Batching}} \\
Effective batch size            & 4 \\
Per-device train/eval batch size     & 1 \\
Gradient accumulation steps     & 4 \\
\midrule
\multicolumn{2}{l}{\textit{Training schedule}} \\
Number of epochs                & 10 \\
\midrule
\multicolumn{2}{l}{\textit{Memory \& efficiency}} \\
Precision                       & bf16 \\
Dataloader workers              & 4 \\
\midrule
\multicolumn{2}{l}{\textit{Checkpointing}} \\
Save total limit                & 1 (best only) \\
Best model metric               & \texttt{eval\_qwk\_5} (higher is better) \\
\bottomrule
\end{tabular}
\end{table}

\subsection{Computational Resources for the Experiments}
\label{appendix:compute}

All fine-tuning experiments were conducted on \textbf{Azure Machine Learning} using a single NVIDIA A100 80\,GB GPU.
Table~\ref{tab:training-times} reports the approximate wall-clock training time for each experiment.

\begin{table}[h]
\centering
\caption{Approximate training times per experiment on a single A100 80\,GB GPU.}
\label{tab:training-times}
\small
\begin{tabular}{lc}
\toprule
\textbf{Experiment} & \textbf{Training Time} \\
\midrule
BLV  & 3\,h\,20\,min \\
DWF-S & 2\,h\,15\,min \\
DWF-N  & 1\,h\,50\,min \\
\bottomrule
\end{tabular}
\end{table}

\clearpage
\section{Further Results}\label{app:results}

This appendix reports additional quantitative and qualitative results that complement the main text. We analyse correlation between alignment meta-metrics and downstream metrics, robustness to holdout set size, detailed behaviour of the best-performing Community Scorers, and extended results for the downstream content filtering application.

\subsection{Correlation between Predictive and Consequential Validity}

\begin{figure}[h!]
    \centering
    \includegraphics[width=1\linewidth]{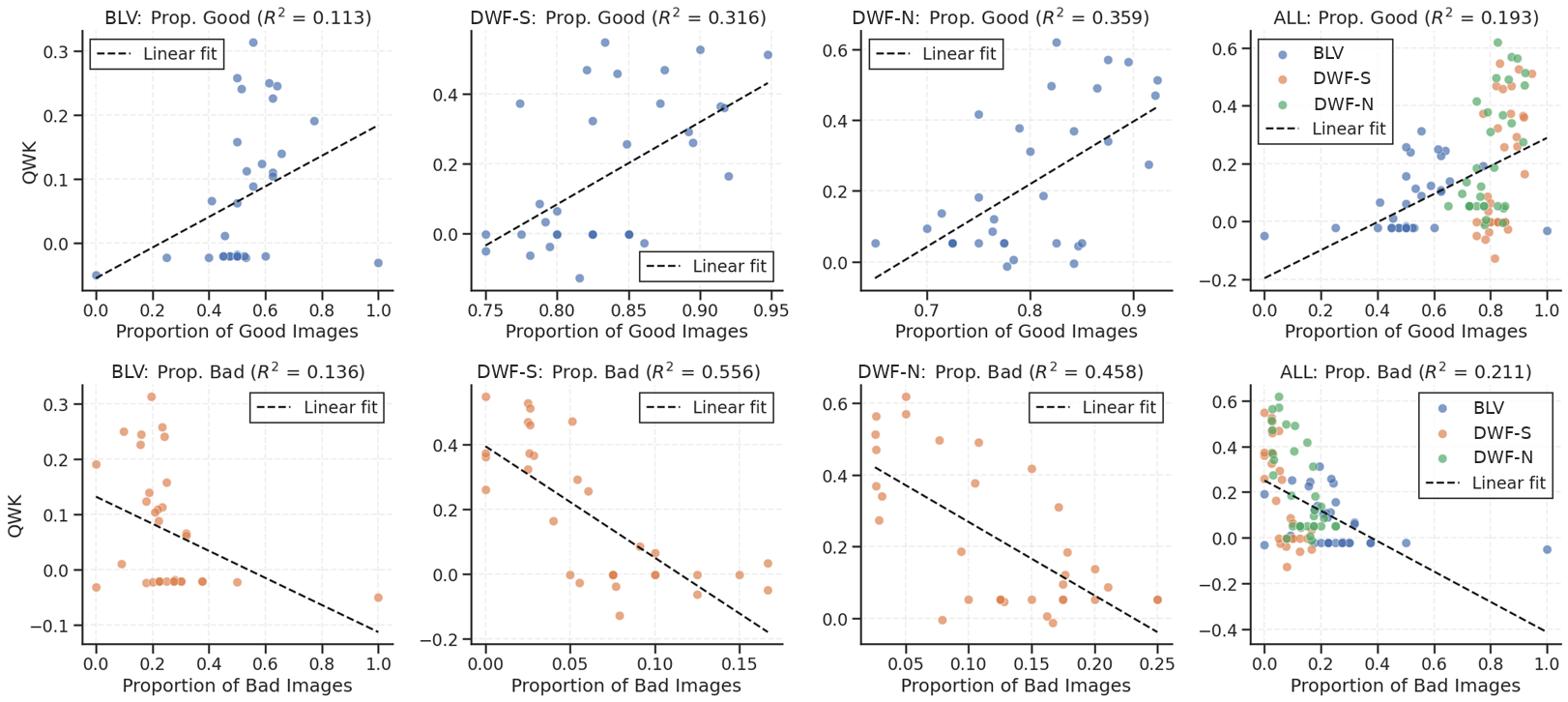}
    \caption{\textbf{Alignment between QWK and downstream quality metrics.} Higher Quadratic Weighted Kappa (QWK) corresponds to a higher fraction of images rated positively and a lower fraction rated negatively by the community.}
    \label{fig:correlation}
\end{figure}

Figure~\ref{fig:correlation} examines the relationship between agreement with human ratings, measured by Quadratic Weighted Kappa (QWK), and downstream quality indicators derived from community preferences. Across communities, higher QWK values are associated with a higher proportion of images rated as good and a lower proportion rated as bad, indicating that stronger ordinal alignment with community judgements translates into improved downstream behaviour. This trend is weaker for BLV, reflecting the greater heterogeneity and complexity of preferences within this community.

Table~\ref{tab:main_result} reports QWK on the held-out set described in the main text. Since QWK requires ordinal predictions, it can only be computed for CAIRE~\citep{yayavaram2025caire} among the baselines. Community Scorers consistently achieve substantially higher agreement with community ratings than generic or non-community-aligned alternatives, underscoring the importance of community-specific~modelling.

\begin{table}[ht]
\centering
\caption{\textbf{Quadratic Weighted Kappa (QWK) on a held-out set by community and method.} QWK is only reported for methods producing ordinal scores. Community Scorers achieve the highest agreement across all communities.}
\vspace{0.2cm}
\label{tab:main_result}
\begin{tabular}{lccc}
\toprule
\textbf{Method} & \textbf{BLV} & \textbf{DWF-N} & \textbf{DWF-S} \\
\midrule
CAIRE & -0.00 & 0.07 & -0.04 \\
Generic Scorer & -0.02 & 0.05 & -0.00 \\
Community Scorer & 0.24 & 0.44 & 0.46 \\
\bottomrule
\end{tabular}
\end{table}




\newpage
\subsection{Analysis of Best Community Scorers}

We further analyse the behaviour of the best-performing Community Scorers using confusion matrices, marginal score distributions, and calibration curves.

\begin{figure}[htbp]
    \centering
    \includegraphics[width=1\linewidth]{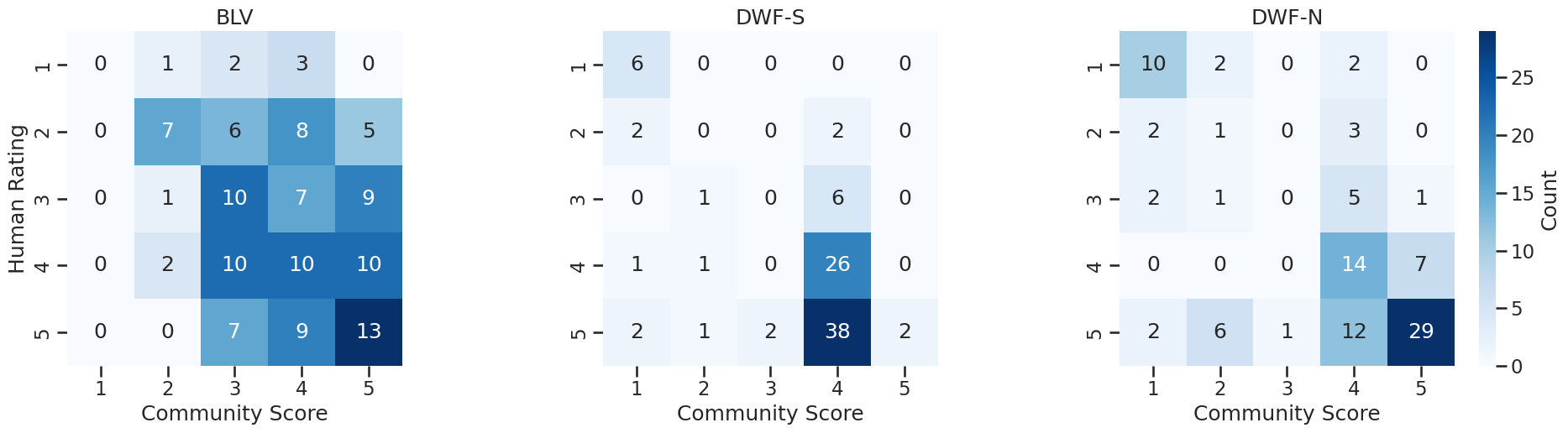}
    \caption{\textbf{Confusion matrices for the best Community Scorers.} Predictions concentrate along the diagonal, with most disagreements limited to neighbouring rating levels.}
    \label{fig:cm}
\end{figure}

Figure~\ref{fig:cm} shows confusion matrices between predicted and human ratings. Most mass lies along the diagonal, indicating strong ordinal alignment, with disagreements primarily occurring between adjacent score levels rather than extreme mismatches.

\begin{figure}[htbp]
    \centering
    \includegraphics[width=1\linewidth]{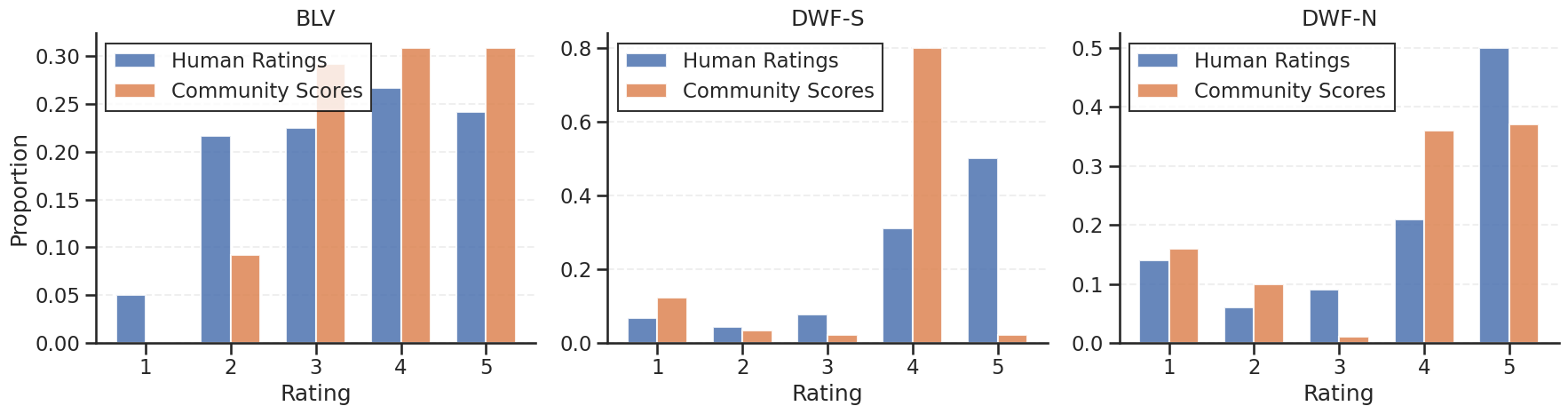}
    \caption{\textbf{Marginal score distributions.} Predicted and human rating distributions are closely aligned, indicating good coverage of the community rating scale.}
    \label{fig:cs_best2}
\end{figure}

Figure~\ref{fig:cs_best2} compares the marginal distributions of predicted scores and human ratings. The close correspondence indicates that Community Scorers capture not only relative ordering but also the overall prevalence of rating levels within each community.

\begin{figure}[htbp]
    \centering
    \includegraphics[width=1\linewidth]{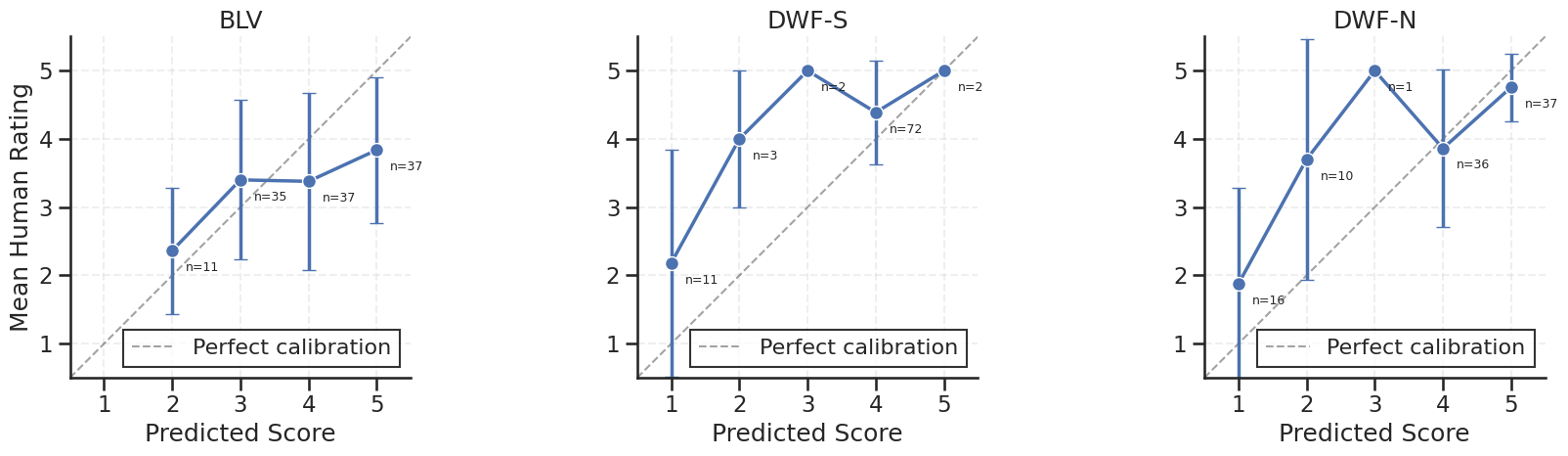}
    \caption{\textbf{Calibration curves for Community Scorers.} Predicted scores closely track empirical human ratings, indicating good calibration across most of the scale.}
    \label{fig:cs_best3}
\end{figure}

Figure~\ref{fig:cs_best3} reports calibration curves, showing that predicted scores are reasonably calibrated with respect to empirical human ratings, with only mild deviations at the extremes of the scale.

\subsection{Results for Content Filtering}\label{app:content_filtering}

\begin{figure}[htbp]
    \centering
    \includegraphics[width=\linewidth]{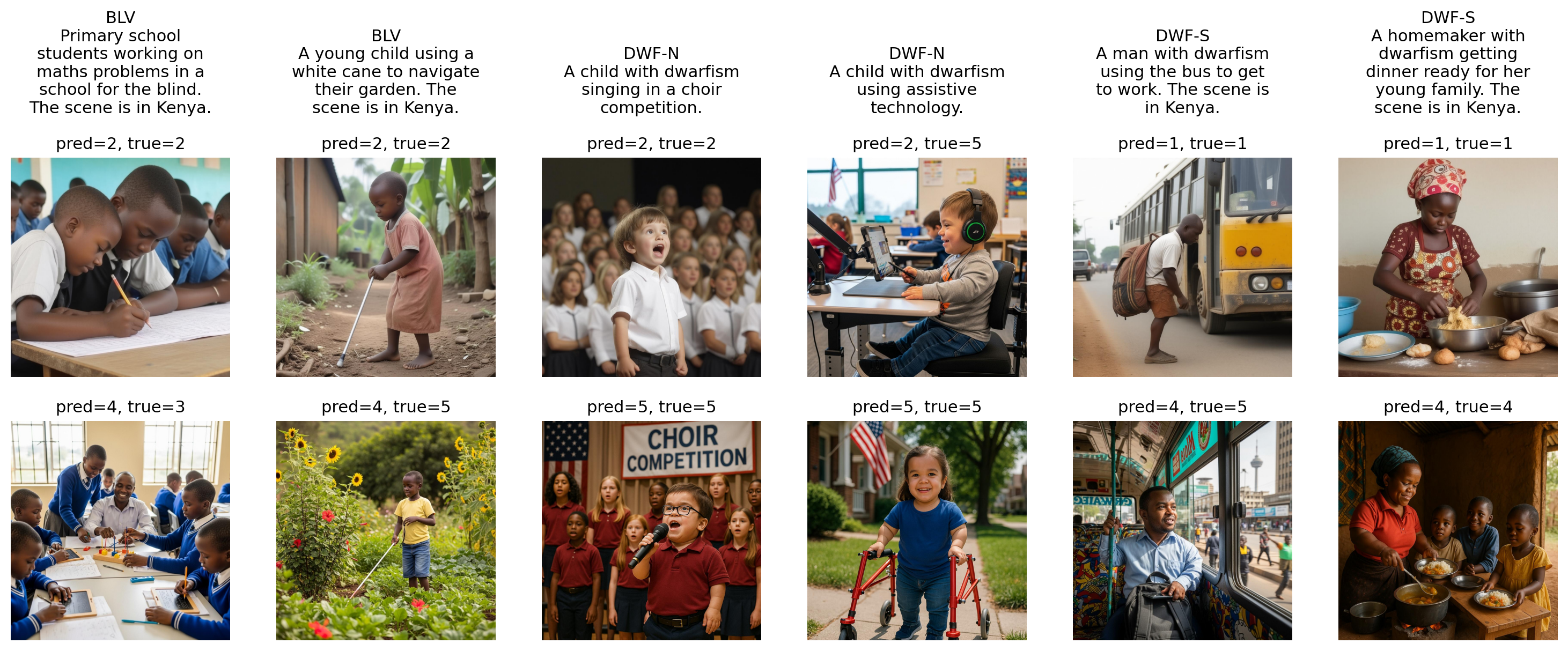}
    \caption{\textbf{Qualitative impact of Community Scorers in content filtering.} Example outputs from the filtering pipeline. Images in the first row are identified as poor representations (low alignment with community ratings) and filtered out, while images in the second row are retained as better aligned with the community's preferred representation.}
    \label{fig:close_loop_qualitative}
\end{figure}

\subsubsection{Threshold Ablation Analysis}

\begin{figure}[htbp]
    \centering
    \includegraphics[width=1\linewidth]{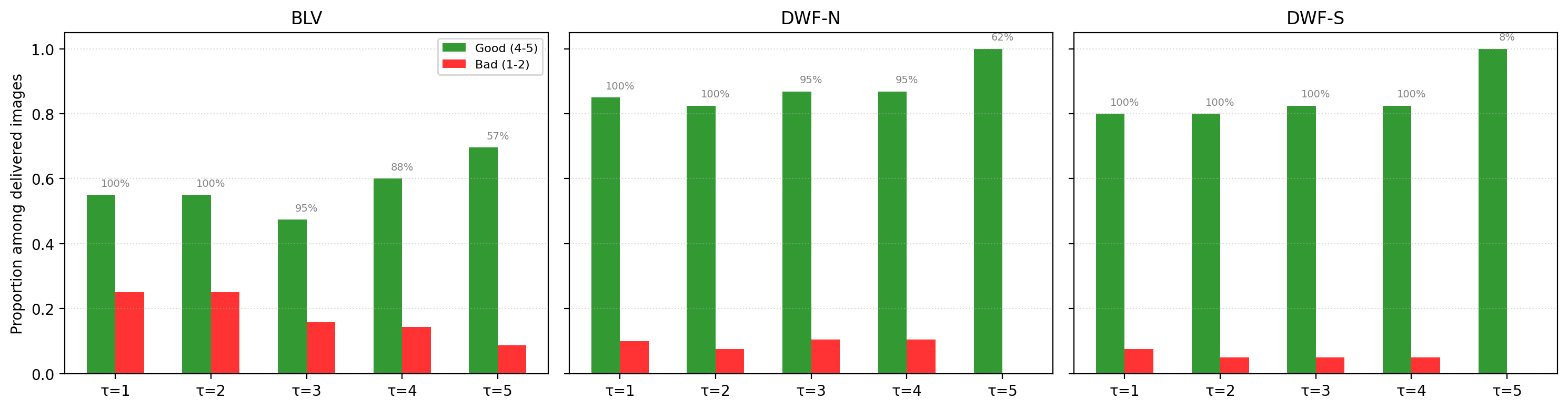}
    \caption{\textbf{Threshold ablation for content filtering.} Intermediate thresholds balance removal of low-rated content with retention of high-quality images.}
    \label{fig:th_ablation}
\end{figure}

Figure~\ref{fig:th_ablation} analyses the effect of the decision threshold used for community-aligned content filtering. Varying the threshold reveals a trade-off between removing low-quality or misaligned content and retaining high-quality outputs, with intermediate thresholds achieving the best balance.

\subsubsection{Qualitative Results}

\begin{figure}[htbp]
    \centering
    \begin{subfigure}[t]{0.85\linewidth}
        \centering
        \includegraphics[width=\linewidth]{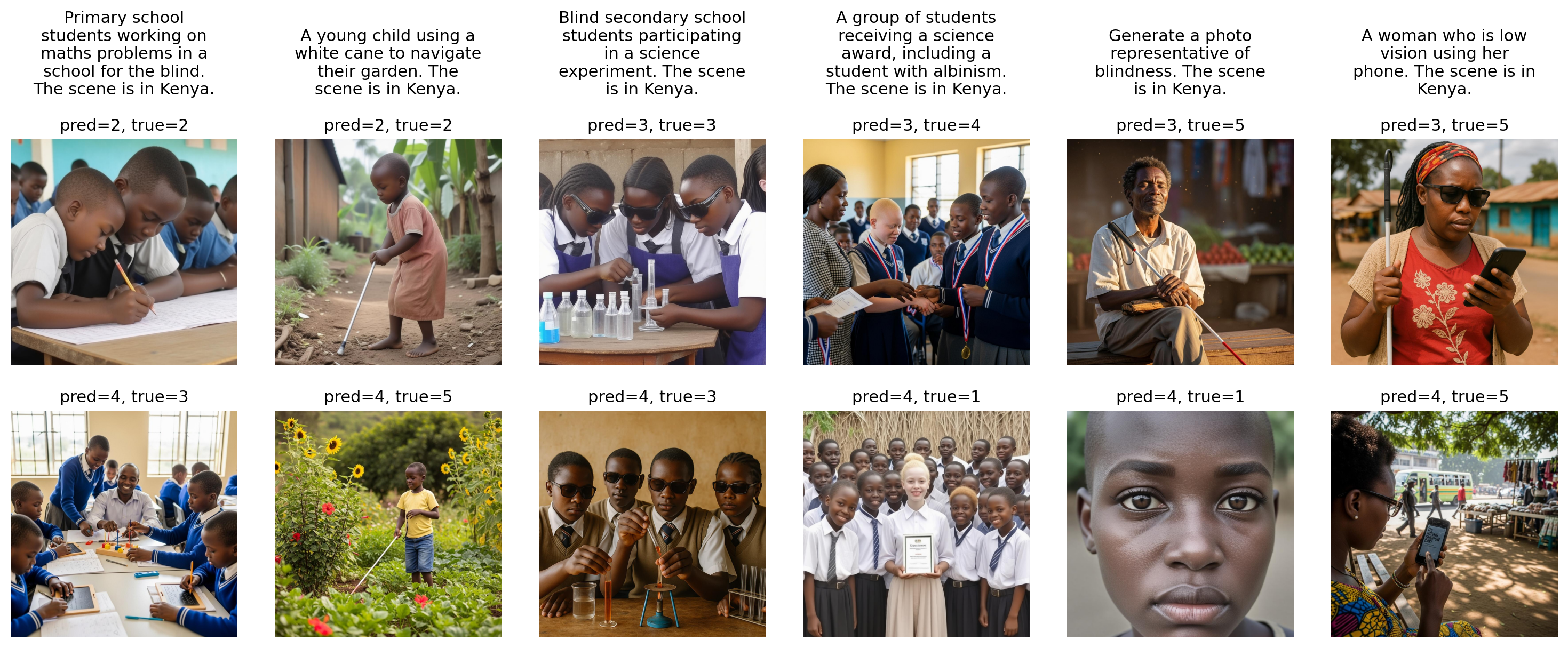}
        \caption{\textbf{BLV}}
        \label{fig:cf_kbt}
    \end{subfigure}


    \begin{subfigure}[t]{0.85\linewidth}
        \centering
        \includegraphics[width=\linewidth]{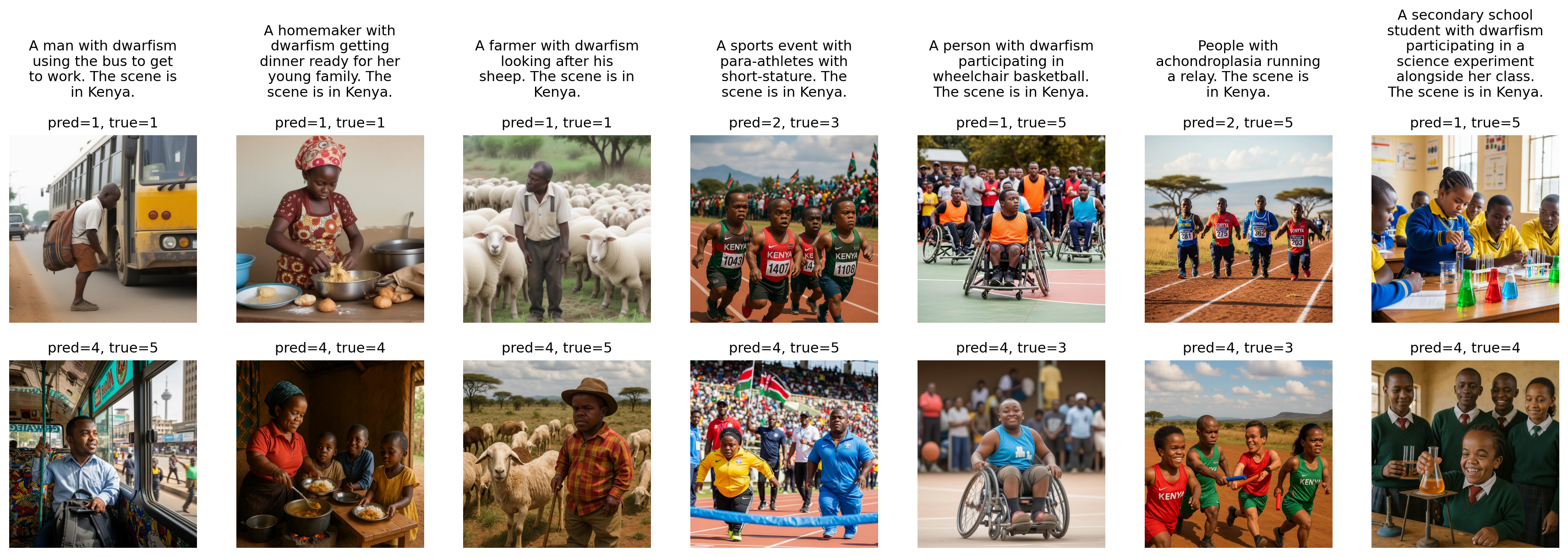}
        \caption{\textbf{DWF-S}}
        \label{fig:cf_sssk}
    \end{subfigure}


    \begin{subfigure}[t]{0.85\linewidth}
        \centering
        \includegraphics[width=\linewidth]{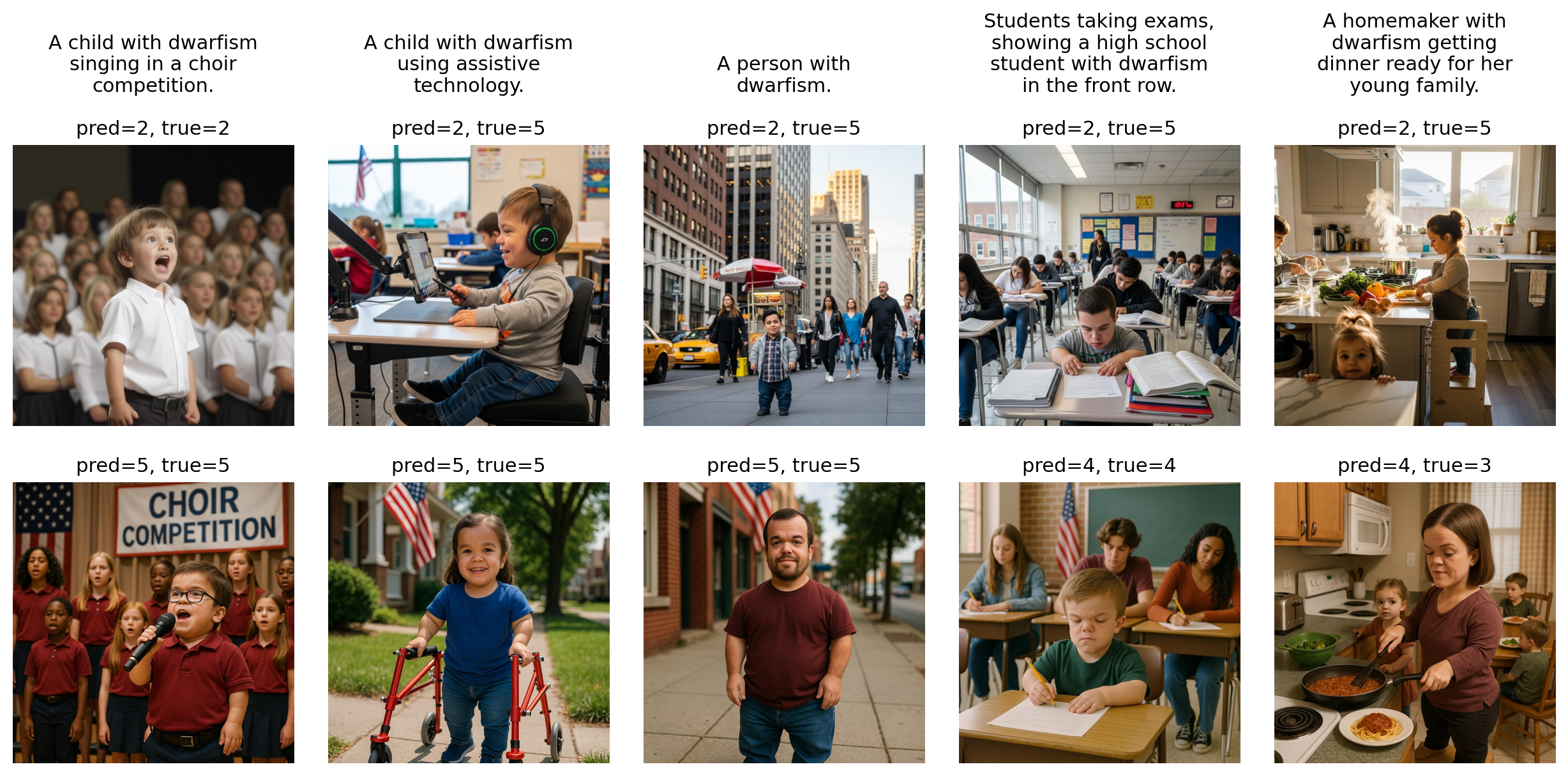}
        \caption{\textbf{DWF-N}}
        \label{fig:cf_lpa}
    \end{subfigure}

    \caption{\textbf{Qualitative results for community-aligned content filtering across communities.} 
    For each community, the Community Scorer filters out images with subtle representational issues or misalignment, yielding final outputs that better reflect community preferences and~norms.}
    \label{fig:cf_qualitative}
\end{figure}

Figures~\ref{fig:cf_kbt}--\ref{fig:cf_lpa} present qualitative examples of content filtering across communities. Each figure contrasts images filtered out (top row) by the Community Scorer with the final retained outputs (bottom row). Filtered images typically exhibit subtle representational issues or mismatches with community expectations, while retained images better align with the prompt and community norms.

\clearpage
\section{Ethical Considerations Statement}\label{app:ethics}

\paragraph{Study Method}
The community organisations that participated in this research were recruited through an international NGO. Research agreements were signed that included financial support for their time and contributions. Research participation by the designated community lead of each organisation was voluntary. Before taking part, interested individuals received information explaining the purpose of the research, what participation involved, the potential risks and benefits, and how their information would be protected. They provided informed consent in writing. The research involving the community-led creation of the Community Preference Dataset was reviewed and approved by the appropriate institutional research ethics review process (IRB ID: 11019) prior to data collection.

This work builds on open-sourced Community Libraries, which contain sensitive, non-anonymous photographic depictions of people (including minors) with disabilities. We ensured that each community organisation has followed appropriate compliance processes, including guardian consent, for those images. All community data was stored within a password-secured, data-safe, and privacy-compliant research environment. The publicly released Community Library datasets on Hugging Face are governed by each community organisation, who control and specify requirements for who can access the data (including access gates and data use licences).

To reduce risks of exposing designated community leads to potentially harmful AI-generated content, we applied a human filtering step. Domain experts in disability removed images that could be considered highly offensive, using the following rubric: 1) any sexualised or violent content; 2) malformed bodies (e.g., non-disability related missing or merged body parts); 3) total erasure of a disability; and 4) demeaning spurious correlations (e.g., portraying someone with dwarfism as an elf). This step did not remove all poor representations, as that can only be determined by the community, but removed images that would be clearly harmful and discouraging to participants.

\paragraph{Community Preference Datasets}
The Community Preference Datasets were developed for research and experimental purposes. They reflect what state-of-the-art AI image generation models could do at the time of creation. These datasets have not been systematically evaluated for sociocultural/ economic/ demographic/ linguistic bias. They should not be used in highly regulated domains where inaccurate or incomplete outputs could suggest actions that lead to injury or negatively impact an individual's legal, financial, or life opportunities. The risks of specific intended downstream uses should also be carefully~considered.

The Community Preference Datasets provide a foundation for advancing both methodological and applied research on community-aligned AI evaluation. The data enables the development and validation of new evaluator models, supporting innovation in how these are constructed, trained, and validated against community ratings. As a resource, these datasets and any evaluator models built from them can lower the barrier to entry for studying participatory, community-grounded evaluation, fostering reproducibility and enabling more systematic progress on representation-aware assessment in generative AI models.

\paragraph{Dataset Release.}
The released Community Preference Datasets provide a foundation for advancing both methodological and applied research on community-aligned evaluation. The data enables the development and validation of new Community Scorers, supporting innovation in how these are constructed, trained, and validated against community ratings. In parallel, the detailed description of Community Scorers and deliberate standard methodological choices offer clear baseline references for future work and can serve as practical proxies in downstream tasks that require similar forms of generalisation as the proposed content filtering downstream task. Together, we lower the significant barrier to entry for studying community-grounded evaluation, fostering reproducibility and enabling more systematic progress on representation-aware assessment in generative~models.

\clearpage
\end{document}